\documentclass[11pt,a4paper]{article}
\usepackage{hyperref}
\pdfoutput=1
\usepackage[pdftex]{graphicx}
\usepackage{pdfpages}
\usepackage{booktabs} 
\usepackage{epstopdf}
\usepackage{graphicx}
\usepackage{paralist}
\usepackage[position=t,singlelinecheck=off]{subfig}
\usepackage{caption}
\renewcommand\thesubfigure{\Alph{subfigure}}
\usepackage{color}
\newtheorem{lemma}{Lemma}

\newtheorem{myDef}{Definition}
 
\usepackage[linesnumbered,ruled]{algorithm2e}
\usepackage{array,color}
\usepackage{listings}
\usepackage{graphicx,subfig}
\usepackage{amssymb,amsfonts,amsmath}
\usepackage[a4paper,left=25mm,right=20mm,top=15mm,bottom=15mm]{geometry}
\usepackage[ruled]{algorithm2e} 
\usepackage{amsmath}
\usepackage{colortbl,booktabs}
\DeclareMathOperator*{\argmax}{arg\,max}
\usepackage{subfig}
\usepackage{threeparttable}
\usepackage{footnote}
\usepackage{balance}
\newcommand{\tabincell}[2]{\begin{tabular}{@{}#1@{}}#2\end{tabular}}
\usepackage{diagbox}
\usepackage{epstopdf}
\usepackage{tablefootnote}
\usepackage{multirow}
\usepackage{enumitem}
\usepackage{amsfonts}
\usepackage[ruled]{algorithm2e} 

\begin{document}
\title{\bf{Bioinspired Bipedal Locomotion Control for Humanoid Robotics Based on EACO}}
\date{}
\author{\sffamily Jingan Yang$^{1,*}$, Yang Peng$^{2}$ \\
    {\sffamily\small $^1$ Institute of AI and Robotics, Hefei University of Technology Hefei 230009, CHINA}\\
    {\sffamily\small $^2$ School of Computer and Information Sciences, Changzhou Inst of Tech, Jiangsu, CHINA}}
\renewcommand{\thefootnote}{\fnsymbol{footnote}}
\footnotetext[1]{Corresponding author}
\maketitle
\begin{abstract}
Over the past decade, substantial progress have been made in the development of humanoid
robots. One of the outstanding challenges in gait optimization of humanoid robots is to find gait
parameters that can optimize a desired performance metric, such as walking speed, robustness, and energy efficiency. The existing gait optimization methods, such as Bayesian optimization~(BO), gradient descent methods~(GDM), genetic algorithms~(GA), and particle swarm optimization~(PSO), are very time-consuming to find the optimal solutions for objective function with multiple local minima. Furthermore, most of global optimization approaches are impractical to apply to the real robots because they require a large number of interactions with robots, such as, for each iteration many sets of parameters from the population in genetic algorithms must be evaluated. Therefore, we must greatly reduce the required number of interactions for finding optimal parameters.

To construct a robot that can walk as efficiently and steadily as humans or other legged animals,
we develop an enhanced elitist-mutated ant colony optimization~(EACO) algorithm with genetic and crossover operators in real-time applications to humanoid robotics or other legged robots. This work presents promoting
global search capability and convergence rate of the EACO applied to humanoid robots in real-time by estimating
the expected convergence rate using Markov chain.
Furthermore, we put a special focus on the EACO algorithm on a wide range of problems, from ACO, real-coded GAs, GAs with neural networks~(NNs),
particle swarm optimization~(PSO) to complex robotics systems including gait synthesis, dynamic modeling of parameterizable trajectories and gait optimization of humanoid robotics.
The experimental results illustrate the capability of this method to discover the premature
convergence probability, tackle successfully inherent stagnation, and promote the convergence rate of the EACO-based humanoid robotics systems and demonstrated the applicability and the effectiveness of our strategy for solving sophisticated optimization tasks. We found reliable and fast walking gaits with a velocity of up to 0.47m/s using the EACO optimization strategy. These findings have significant implications for understanding and tackling inherent stagnation and poor convergence rate of the EACO and provide new insight into the genetic architectures and control optimization of humanoid robotics. %
\end{abstract}
%
%
\section{INTRODUCTION}
In order to realize gait optimization of the simulated and physical humanoid robots based on the EACO, this section provides an overview of {\textit{the state-of-the-art}} from contributing fields such as the EACO, humanoid robotics, gait optimization, and human-robot interaction~\cite{yang12}. Gait generation of humanoid robots is a very challenging task, because their bipedal locomotion systems have many Degrees of Freedom (DOFs) with unstable, nonlinear and underactuated dynamics, and gait optimization determines the optimal position, velocity, and acceleration for each DOF of humanoid robots at any moment and their movement quality. Building robots that can not only learn from the unknown environments, but judge the risks that they are taking is key to building smarter robots. Therefore, humanoid robots represent one of the ultimate goals of robotics which synthesize advances from many disciplines~\cite{fukuda01}. We have found that several key technologies are of importance for solving sophisticated optimization tasks for the next generation robots.

One of the ultimate goals of humanoid gait optimization based on EACO is to achieve the stability and fast walking of humanoid robotics.
However, in many optimization algorithms, a large amount of objective functions should be evaluated before converging to an optimal solution \cite{jin01,igel01,branke01}. Actually, we can only evaluate very few objective functions in practical applications. We found that a promising strategy to solve this problem seems to just evaluate the solution quality of the individuals.

To solve the optimization problems in real-world applications, theoretical research and performance analysis of
the ACO convergence properties have become a critical issue for designing
and developing the algorithms. Furthermore, theoretical evidence for the convergent
properties of a variety of EACO algorithm applied to bipedal locomotion control should be validated further.
Niehaus and Yun {\textit{et al.}} used particle swarm optimization to speed up the walking capabilities
of humanoid robotics \cite{niehaus02,yun01}. They considered 14 parameters and optimized the parameters for different
walking directions to be allowed for omnidirectional walking. Hemker {\textit{et al.}} used
sequential surrogate optimization to search optimum solutions of the selected gait parameters for forward walking \cite{hemker02}.
Geng {\textit{et al.}} proposed reinforcement learning approach based on a policy gradient to optimize the
parameters of a neuronal controller for a bipedal robot {\cite{geng02}}. Garro {\textit{et al.}} optimized the search of the
shortest path of the mobile robots using evolution of some parameters of the ACO algorithm with a genetic
algorithm~(ACO-GA) \cite{garro02}. Yang {\textit{et al.}} and Islam proposed an improved
ACO algorithm with genetic operator which reconstructs solutions to COPs based on the techniques inspired by
natural evolution including selection, mutation, and crossover {\cite{yang02,islam01}}. Different
optimization approaches pursued by scientists to make progress toward robot bipedalism are validated {\cite{pavlus01}}.

The ACO algorithm based on the cooperative behavior of real ant colonies is one of many promising methods in machine learning and
combinatorial optimization~{\cite{dorigo03}} and is used widely
to solve the complex optimization problems. Actually, each foraging process of ant colonies is
also a learning process. Furthermore, ant colonies navigate their foraging behaviors using their own intelligence and
experience~(e.g. pheromone trails)~{\cite{li03,bonabeau01}}. To strengthen convergent properties
of the EACO algorithms for solving complex combinatorial optimization problems (COPs), we consider the
combinatorial optimization model $({\bf{S}},\Omega,f)$, where ${\bf{S}}$ represents the discrete solution
space, an objective function $f(s)$ is minimized when ${\bf{S}}{\to}\mathbb{R}_0^+$; $\Omega$ is the constraint set
which defines feasible solution set. For any $s{\in}{\bf{S}}$, the values of $f(s)$ can be calculated.
The combinatorial optimization problems solved by the EACO are regarded as search $s^*{\in}{\bf{S}}$
which satisfies $f(s^*){\leq}f(s)$, ${\forall}s{\in}{\bf{S}}$.
\textnormal{\small For walking shortest path planning and gait optimization of a humanoid robot,
suppose $C=\{c_1, c_2, \cdots, c_n\}$ is a set of $n$ nodes, distance between any two nodes is denoted
as $d(c_i, c_j)$, $i,j\in R$, and abbreviated to $d_{ij}$ and satisfies $d_{ij}=d_{ji}$. The $d_{ij}$
can be formulated as follows:}
\begin{eqnarray}
d_{ij}=\sqrt{(x_i-x_j)^2+(y_i-y_j)^2}
\label{eqn:01}
\end{eqnarray}
\begin{myDef}
\textnormal{\small Transition probability $p_{ij}(k,t)$ for ant $k$ at the current node $i$ to select next node $j$ at iteration $t$ is calculated by the following equation:}
\begin{eqnarray}
p_{ij}(k,t)=\left\{\begin{array}{ll}
\frac{[\tau_{ij}(t)]^\alpha[\eta_{ij}]^{\beta(i,t)}}{\sum\limits_{k{\notin}\mathrm{Tabu}_k}[\tau_{ik}(t)]^\alpha[\eta_{ik}]^{\beta(i,t)}} &  ~\mathrm{if}~j{\in}\mathrm{allowed}_{\it k}   \\
\qquad\quad 0    & ~\mathrm{otherwise}
\end{array} \right.
\label{eqn:02}
\end{eqnarray}
\end{myDef}
%
\noindent where $\mathrm{allowed_{\it k}}$=$\{C-\mathrm{Tabu}_k\}$---next nodes which the $k^{th}$ ant is allowed to choose at iteration $t$. $\mathrm{Tabu}_k$ is a ${\color{blue}{\mathrm{taboo~list}}}$ which contains all edges visited by ant~$k$. At this iteration,
the $k^{\mathrm{th}}$ ant does not visit again these nodes. $\tau_{i,j}(t)$ is called the concentration of pheromone
related to edge $(i,j)$ at iteration $t$, $\eta_{i,j}$ represents the desirability of edge $(i,j)$, $\alpha$ and $\beta$ represent two
decision-making policy control parameters. $\alpha$ is used to control the relative weight of the pheromone trail in the decision making process,
and the role of $\beta$ is to control the relative importance of the desirability and the influence of $\eta_{i,j}$,
also referred to as the heuristic sensitivity parameter, in the decision-making process, as shown in~Eq.~[{\ref{eqn:02}}].
The influence of $\alpha$ and $\beta$ upon the performance of the EACO algorithms is explained in detail as: ($i$) If $\alpha=0$,
the cheapest options are more likely to be selected which may result in a
classical stochastic greedy algorithm; ($ii$) On the contrary if $\beta=0$, only pheromone increment occurs, ACO performance is rapidly poor
and will lead to the premature convergence of the algorithm to a suboptimal solution~{\cite{zecchin01,dorigo01,afshar03}}.

The $\eta_{ij}$ is the heuristic desirability defined by the reciprocal of distance $d(c_i, c_j)$, $\eta_{ij}=\frac{1}{d(c_i,c_j)}$,
namely, $\eta_{ij}=\frac{1}{d_{ij}}$,
where $d_{ij}$ is the Euclidean distance from node $i$ to node $j$~which can be calculated by~Eq.~[{\ref{eqn:01}}].
The $\eta_{ij}$ is computed once the algorithm operates and will be not changed during the whole computation.
$\tau_{ij}$ is the intensity of pheromone trail deposited on edge $(i,j)$.

The execution process of the ant colony algorithm may be regarded as an ant's stochastic walking on oriented graph $G=(C,L,P)$.
The discrete set $C$ is regarded as the nodes of the graph $G$; $L$ is the edge set, which describes the
link situation of the edges in the graph $G$; ${\bf}P$ is the pheromone vector and $\tau_{ij}$ corresponds to the pheromone on edge $(i,j)$ from $c_i$ to $c_j$.

In this work, we aim at constructing humanoid robots that can walk as efficiently and steadily as humans which is a central challenge in the robotics areas. Building on former findings on deterministic chaotic activities of ant colonies, we developed a novel framework of the EACO algorithms for the locomotion control and gait optimization of humanoid robotics. Our strategies for enhancing global search capability and the convergence rate of the EACO in real-time applications to humanoid robotics is proven to faster converge and can reduce the premature convergence probability of the EACO systems. Moreover, we demonstrate that the enhanced EACO algorithm decrease the premature convergence probability, promote the convergence rate of the EACO system, and prevent humanoid robotics systems from getting trapped into local optima. In particular, the EACO can efficiently find gait parameters that optimize the desired performances for bipedal locomotion. These findings have significant implications for understanding and tackling inherent stagnation and poor convergence rate of the EACO and realizing walking trajectory planning and gait optimization of humanoid robotics with a faster convergence rate. In the experiments,
we found reliable and fast walking gaits with a velocity of up to 0.48m/s.

\vspace*{-0.1cm}
\section{RESULTS}
\subsection{EACO-Based Robotics Control}
In the experiments that were conducted following the EACO algorithms, we complete bioinspired design as applied to robotics aiming at implementing naturally occurring features and validate the EACO algorithms applied to bipedal locomotion control of humanoid robotics~(Video~{\color{red}II} and Fig.~{{\ref{efig10}}}).
The EACO algorithms applied to path planning and gait optimization of humanoid robots are investigated and the proposed strategy is proven
to faster converge. Moreover, global search capability and convergence rate of the EACO in real-time
applications to humanoid robots are promoted by estimating the expected convergence rate using Markov
chain~{\cite{ding01}}. The experiments on the simulated and real robots demonstrate that performances of the proposed algorithm are superior to that of the GAs and standard ACO. More recently, we have developed several techniques to optimize the
behavior or the motions of the humanoid robots and optimized successfully walking stability, walking speed and parameterized gait
of the humanoid robots (Fig.~{{\ref{efig09}}} and Fig.~{{\ref{efig10}}}).

We also use mutation operator and self-adaptive approach for enhancing the algorithm capability
to escape from local minima. This could solve not only combinatorial optimization problems, but also could preclude
local minima and promote convergence speed for path planning and gait optimization of the simulated humanoid robots {\cite{chen01}}.
In this article, we deal with convergence property of the EACO algorithms based on the biological inspiration which can
be transferred into a strategy for solving discrete optimization problems, and present some variants of best-performing
EACO for solving the optimization problems {\cite{birattari01}}. The theoretical results has proved that ant
colony optimization algorithms can be used to solve complex combinatorial optimization problems, for example,
motion trajectory planning and gait optimization of physical humanoid robotics.

For self-adaptive crossover and mutation based on the solution quality, the EACO algorithms can update the pheromone intensity
on edge ($i,j$) based on the quality of solutions constructed by ants, that is, the higher quality solutions will contribute more
pheromone, the worse solutions contribute less pheromone. Furthermore, the population diversity is calculated by using genotype
method and quality approximation is denoted by using weighted average. As we can see self-adaptive strategy is able to perform
very much both crossovers, such as, edge-recombination crossover and partial matching crossover. Average solution qualities
obtained by edge-recombination crossover strategy is superior to average solution qualities obtained using partial matching crossover
strategy {\cite{aine01}}. The results of adaptive parameter control strategies of the enhanced EACO applied to COPs are illustrated in Figs.~{{\ref{efig01}}}({\bf A}) and {{\ref{efig01}}}({\bf B}).
\subsection{Strategies to Enhance Global Convergence Rate}
To enhance global convergence rate of the EACO, suppose we have a set $|D[k]|$$D[k]=\sum\limits_{{i,j}{\in}C}d_{ij}[k],~k=|1,2,\cdots,n|$
%
\begin{align}
d(t)_{\mathrm{min}}&=\mathrm{min}[D[1],D[2],\cdots,D[n].  \\
d(t)_{\mathrm{aver}}&=\frac{[D[1]+D[2]+,\cdots,+D[n]}{n}.
\label{eqn:03}
\end{align}
where:
\begin{itemize}
\item $d_{ij}[k]$ is distance of its tour that ant $k$ constructs from node $i$ to node $j$ at each iteration;
\item $d(t)_{\mathrm{min}}$ represents the length of the shortest path on edge~($i,j$) that the $k^{th}$ ant traveled at each iteration $t$;
\item $d(t)_{\mathrm{aver}}$ denotes a mean path length of their tours that $n$ ants traveled at each iteration $t$.
\end{itemize}
Only while $d(t)_{\mathrm{min}}$$<$$d(t-1)_{\mathrm{min}}$ and $D(k)$$<$$d_{\mathrm{aver}}(t)$,
the $k^{th}$ ant is able to compute $\Delta\tau^k_{ij}(t)$ according to~the following equation
\begin{eqnarray}
\tau_{ij}(t+1)=(1-\alpha)\tau_{ij}(t)+\alpha\Delta\tau_{ij}{(t)}.
\label{eqn:04}
\end{eqnarray}
namely, only elitist individuals (ants) that pass the shortest paths in current cycle are allowed
to modify $\tau_{ij}(t)$.
\subsection{Adaptive Control of Pheromone Trails}
Ant behavior is in fact the inspiration for the metaheuristic optimization techniques which can be
used to control the pheromone density deposited on the path that an ant traverses based on
the solution quality. At the same time, we also use an adaptive pheromone
update rule to update pheromone trails which is formulated as:
\begin{equation}
\rho_n=L_n^{-1}/(L_n^{-1}+L_{pn}^{-1})
\label{eqn:05}
\end{equation}
where:
\begin{itemize}
\item $L_n$ represents the solution $S_n$ constructed by the $n^{th}$ ant at each iteration;
\item $L_{pn}$ denotes the solution $S_{pn}$ obtained by ant $n$ using pheromone matrix, as shown in the following equation.
\end{itemize}
\begin{eqnarray}
s={\mathop{\argmax}_{u{\in}J_n(r)}\{\tau(r,u)\}}
\label{eqn:06}
\vspace*{-0.4cm}
\end{eqnarray}
where:
\begin{itemize}
\item $s$ represents the node which is chosen as next node to node $r$ for $\forall(r,s){\in}S_{pn}$;
\item $S_{pn}$$\raisebox{0.3mm}{--}$the solution produced by ant $n$ based on pheromone matrix;
\item $J_n(r)$ represents the set of the nodes the ant $n$ at the current node $r$ will travel.
\end{itemize}
The proposed adaptive update strategies include: First, we use chaos perturbation to improve nodes
selection strategy and dynamically adjust the evaporation coefficient for improving the global search
capability based on the pheromone intensity. Second, adaptively update the pheromone density based on the
solution quality. The convergence performance can be significantly improved by this method. In addition,
the idea of the ACO metaheuristic method also is used to control pheromone update, namely,
the optimal solutions should deposit more pheromone on edge~$(i,j)$ which ants travelled, and
the poor solutions deposit less pheromone on edges.

The ant colony optimization algorithm usually converges to local optima and the convergence speed is slower. To enhance
global searching capability and performance of the EACO algorithm, first, we choose to use the elitist strategy for reserving
its optimum solution at each iteration. Then, only one solution, namely, the iteration-best solution or the global best solution,
is selected to deposit pheromone trails and bound all deposited pheromone values within $[\tau_{\mathrm{min}},\tau_{\mathrm{max}}]$
in all optimal paths searching at any moment, $t_{ij}(t){\subset}[t_{\mathrm{min}},~t_{\mathrm{max}}]$~{\cite{stutzl02}}.
$\tau_{\mathrm{min}}$ may preclude effectively the algorithm stagnation, and $\tau_{\mathrm{max}}$ can avoid too much pheromone on some
path than that on other paths so that all ants pass on the same
path which limited the algorithm proliferation. Thus, an optimal path is kept at each iteration. Only the ant travelling
the shortest path is authorized to modify $\tau_{ij}(t)$.
\begin{equation}
\tau_{ij}(t+1)=\left\{\begin{array}{ll} \tau_{\mathrm{min}}  & \textrm{if}~\tau_{ij}(t)<\tau_{\mathrm{min}} \\
\tau_{ij}(t)  & \textrm{if}~\tau_{\mathrm{min}}<\tau_{ij}(t)<\tau_{\mathrm{max}}    \\
\tau_{\mathrm{max}}    & \textrm{if}~\tau_{ij}(t)>\tau_{\mathrm{max}}
\end{array} \right.
\label{eqn:07}
\end{equation}

The performance of the proposed algorithms is analyzed and validated by real data. We have developed several algorithms
used to solve COPs in robotics and intelligent control areas. Of them, ant colony algorithm~(ACO){\cite{yang01}}, particle swarm
optimization~(PSO) {\cite{niehaus02}}, genetic algorithm~(GA) and simulated annealing~(SA){\cite{pereira01}},
Tabu search~{\color{red}\footnote{Tabu search is a metaheuristic algorithm that can be used for solving combinatorial optimization problems {\color{blue}\cite{blum01}}}}, are major methods.

{\bf{Table}}~{\ref{tab01}} illustrates the performance comparison results of the EACO, the ACO, the GA, and the PSO.
In our algorithm, mutation and crossover operators are used to make the EACO algorithm escape from local optima.
This algorithm can converge to the optimal solution by optimizing the sub-optimal solutions based on Elitist~strategy. We conclude
from our experimental results that the EACO can solve not only combinatorial optimization problems, but is robust,
able to prevent local optima, and has faster convergence rates than convergence rates of the ACO, the PSO, the GA and the SA.
The SA is a simple Monte Carlo approach for simulating the behavior of the system to achieve thermal equilibrium at
a given temperature in a given annealing schedule {\cite{bella01,ma01}}. We have applied this analogy to tackling
the combinatorial optimization problems and compared the EACO with the GA and the SA methods in solving the
complex combinatorial optimization problems.

{\textbf{Table}}~{\ref{tab02}} illustrated we could find fast the optimal solutions using the EACO algorithm, and these
solutions usually can stabilize in the optimal solutions or sub-optimal solutions after achieving
the optimal solutions. The experimental results and the comparison of calculating $G_1$-$G_4$ using the EACO, the GA, and the SA are illustrated in {\textbf{Table}} {\ref{tab02}}. From the results in {\textbf{Table}}~{\ref{tab02}}, we conclude that
the optimal solutions of $G_1, G_2, G_3$, and $G_4$ calculated by the EACO
most approach to their theoretical optimal solutions {\color{blue}{($-$15.012, 7050.331, 680.538, and 0.0562)}}
separately~{\cite{yang01}}. In our experiments, the number of the initial
solutions: $N$=20, the individual number: $M$=40, and we performed 40 tests on every problem with the
different parameter values using EACO repeatedly, and improved optimal solution precision
by averaging the testing values of repeated experiments, such as $G_1$-$G_4$ proposed by Michalewicz {\cite{michalewicz01}}.

\begin{figure*}[!t]
\centering
\subfloat[]{\includegraphics[width=8.0cm,height=4.0cm]{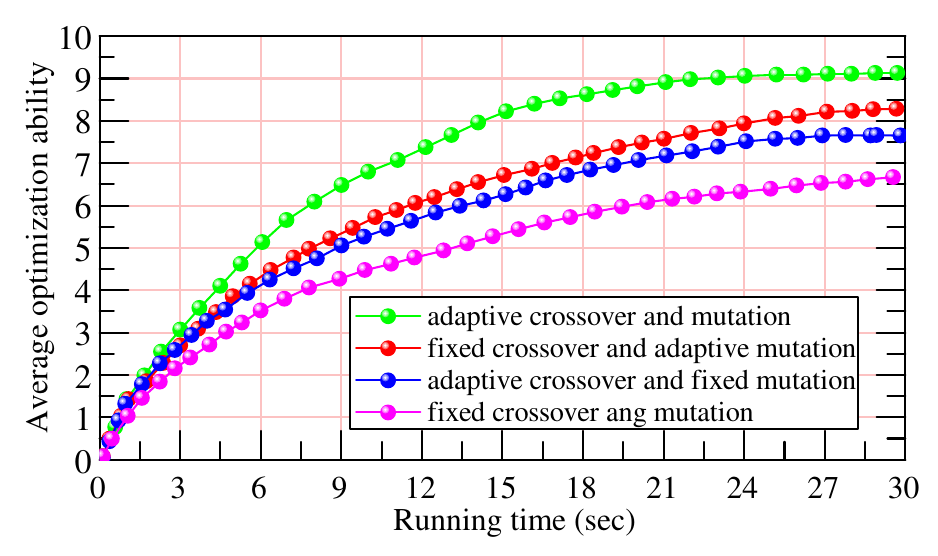}}   \hspace{1pt}
\subfloat[]{\includegraphics[width=8.0cm,height=4.0cm]{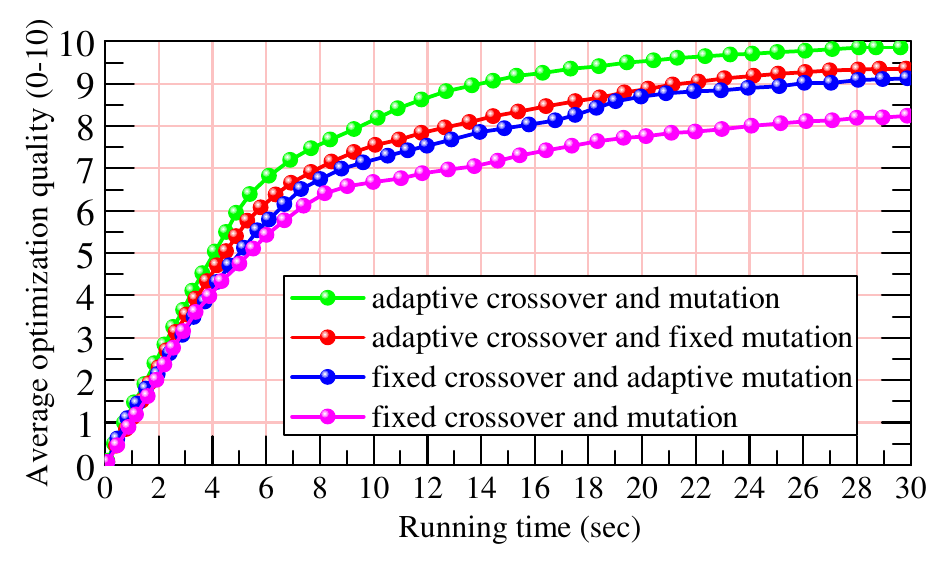}}   \hspace{1pt}    \\
\subfloat[]{\includegraphics[width=8.0cm,height=4.0cm]{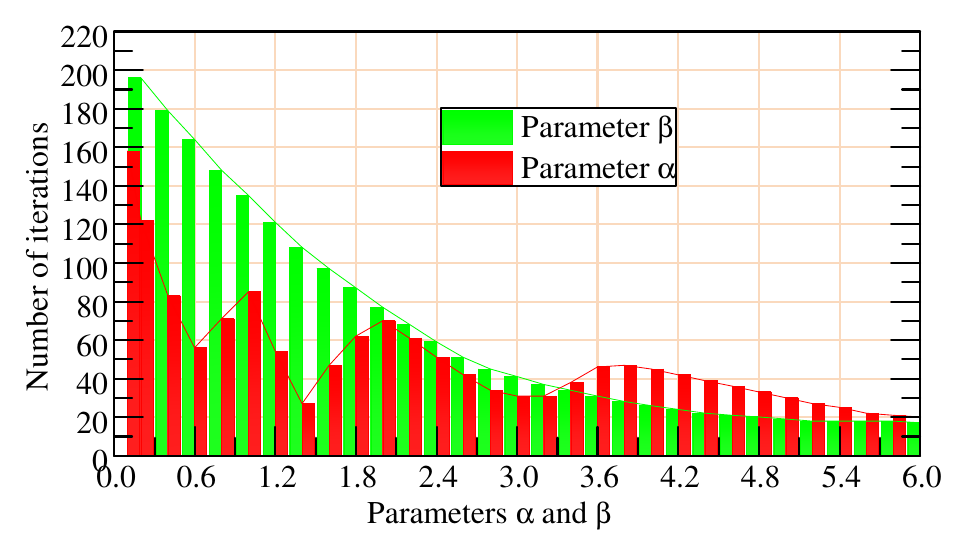}}   \hspace{1pt}
\subfloat[]{\includegraphics[width=8.0cm,height=4.0cm]{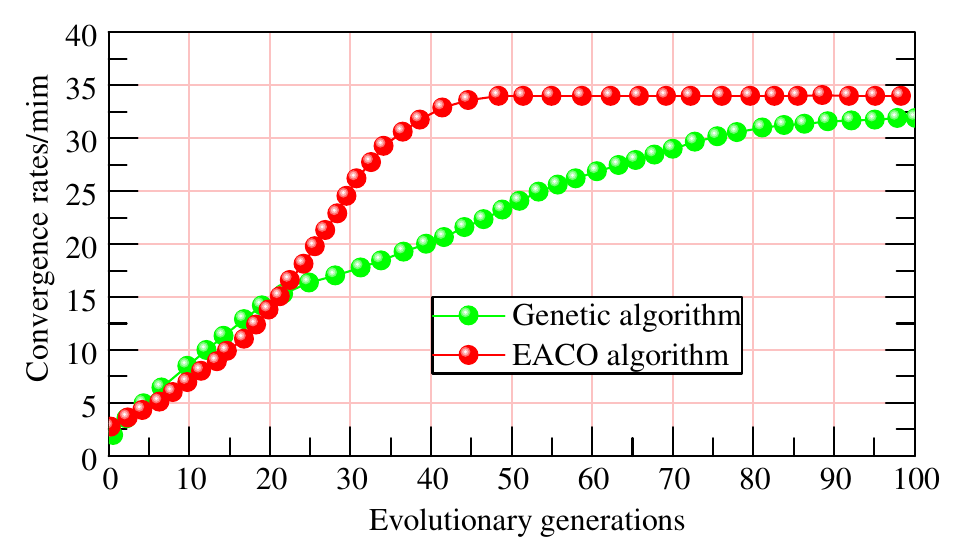}}   \hspace{1pt}
\vspace*{-0.2in}
\caption{\small{Adaptive parameter setup strategies of the EACO applied to COPs and comparison between convergence processes of the EACO and the GA.} (A) the population diversity is calculated using genotype method for edge-recombination crossover strategy. (B) The population diversity is computed using genotype method for partial matching crossover strategy. (C) Effect of {\color{blue}\bf $\alpha$} and {\color{blue}\bf $\beta$} values on convergence rate. (D) Comparison between the convergence processes of the EACO and the GA. Obviously, the larger the parameter {\color{blue}\bf $\beta$}, the faster the convergence rate of EACO algorithm. But if {\color{blue} \bf$\beta$} is extremely large, the convergence property of the EACO tends to get worse. In fact, the parameter {\color{blue}\bf $\beta$} is strongly associated with the parameter {\color{blue}\bf $\alpha$}.}%
\label{efig01}
\end{figure*}

{\textbf{Table}}~{\ref{tab03}} shows effect of the different parameter values on the experimental results of $G_1$ {\cite{zhan01}}.
Therefore we observed that the smallest iteration number is ({\color{red}393.6}) and the optimal
solution is {\color{red}$-14.952$}, when $q_0=0.8$, $\rho=0.3$,
$p_{\rm{crossover}}=0.5$, $p_{\rm{mutation}}=0.6$. Where $q_0$ is a parameter~$(0{\leqq}q_0{\leqq}1)$ which is applied to control
to exploration level executed by the ants. Therefore we can develop the performance of the proposed algorithm and promote convergence
rate by optimizing parameters using adaptive control of pheromone trails and differential evolutionary optimizer
(see Fig.~{\color{red}{\ref{efig09}}} and fig.~{\color{red}{\textrm S4}}) {\color{blue}\cite{qin01,schmidt01}}. Fig.~{\color{red}{\ref{efig01}}} illustrates the results for adaptive parameter control strategies of the enhanced EACO applied to COPs and comparison between convergence processes of genetic algorithm and the EACO. The population diversity is calculated by using genotype method for edge-recombination crossover strategy, as shown in Fig.~{\color{red}{\ref{efig01}}}({\bf A}). The population diversity computed by using genotype method for partial matching crossover strategy is depicted in Fig.~{\color{red}{\ref{efig01}}}({\bf B}).  Fig.~{\color{red}{\ref{efig01}}}({\bf C}) validates the influence of $\alpha$ and $\beta$ values on convergence rates. Comparison between the convergence processes of genetic algorithm and the EACO is depicted in Fig.~{{\ref{efig01}}}({\bf D}). Obviously, the larger the parameter $\beta$, the faster the convergence rate of EACO algorithm. But if $\beta$ is extremely large, the convergence property of the EACO tends to get worse. In fact, the parameter $\beta$ is strongly associated with the parameter $\alpha$.
{\textbf{Table}}~{\ref{tab04}} illustrates the experiential values of the parameters setup of the EACO algorithms applied to path planning and gait optimization for humanoid robots.
\subsection{The Optimized Results of Humanoid Robotics}
\begin{figure*}[!t]
\renewcommand\thesubfigure{\Alph{subfigure}}
\centering
\subfloat[]{\includegraphics[width=16.0cm,height=4.8cm]{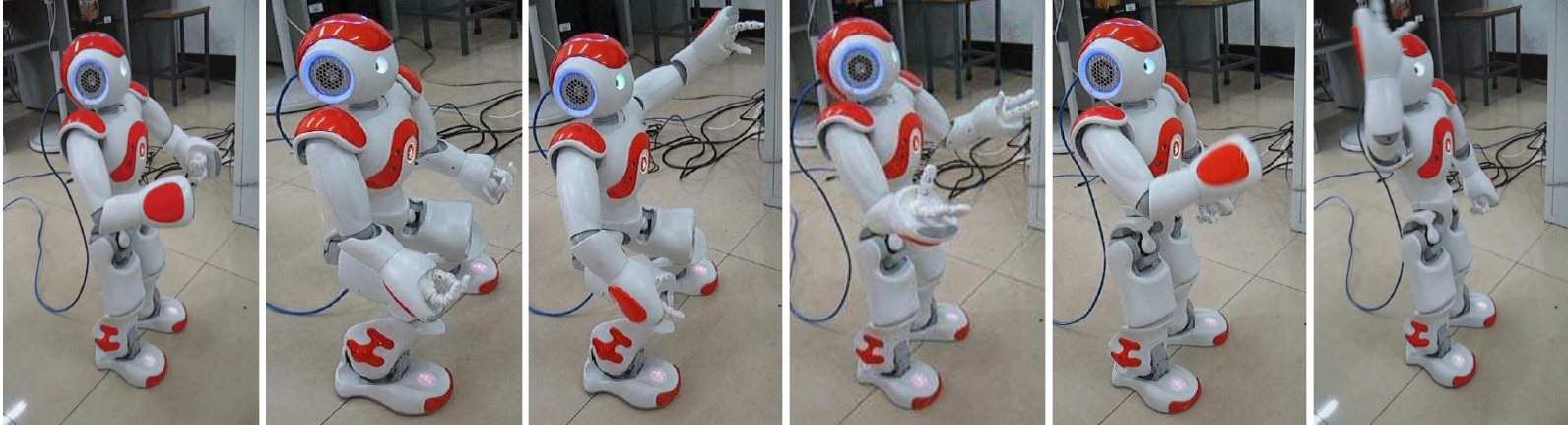}} \hspace{0pt}  \\
\subfloat[]{\includegraphics[width=16.6cm,height=8.8cm]{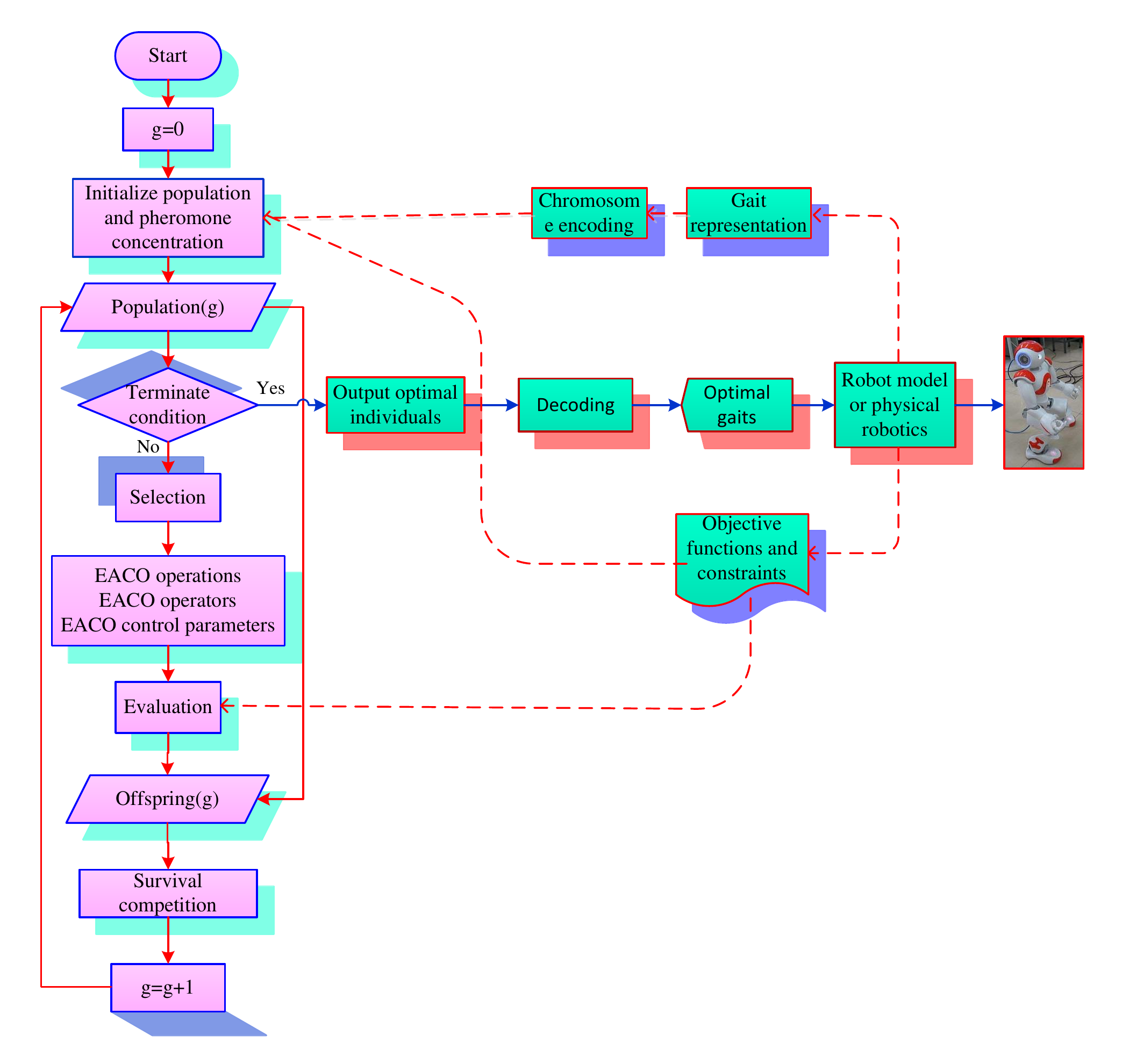}} \hspace{0pt}
\caption{{\bf{Omnidirectional dynamic gait optimization of humanoid robots.}} \small{(A) The experiments towards human-like walking have been
performed on the NAO humanoid robot with its 22 degrees of freedom using the EACO with the parameters described in
Table {\color{red}I}. (B) Flow chart of parameter optimization to obtain optimal walking parameters for gait optimization of humanoid robotics based on EACO.}}%
\label{efig02}
\end{figure*}

\begin{figure*}[!t]
\centering
\subfloat[]{\includegraphics[width=8.0cm,height=4.2cm]{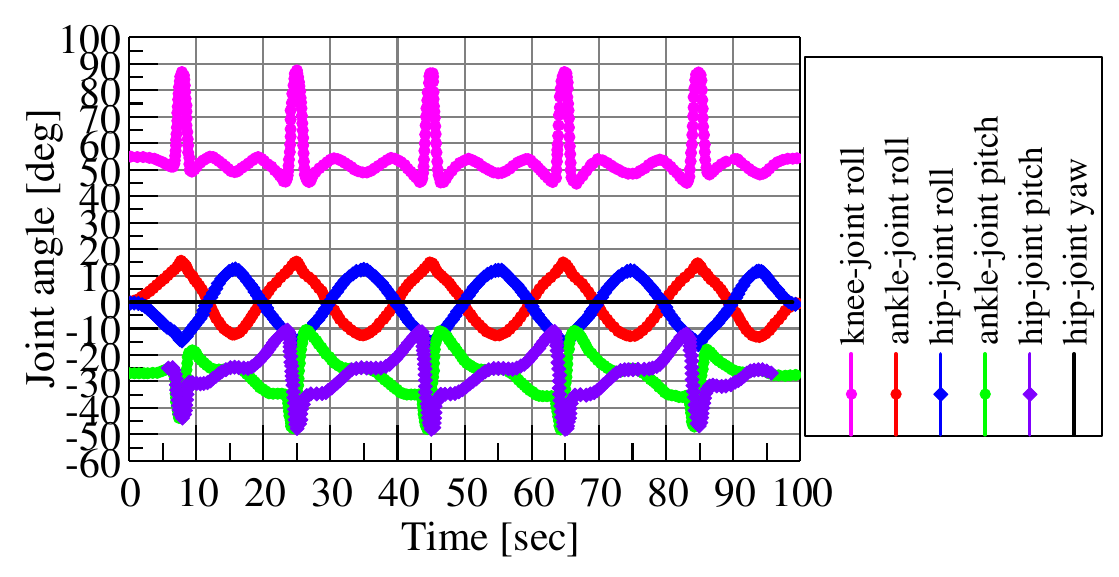}} \hspace{0pt}       %
\subfloat[]{\includegraphics[width=8.0cm,height=4.2cm]{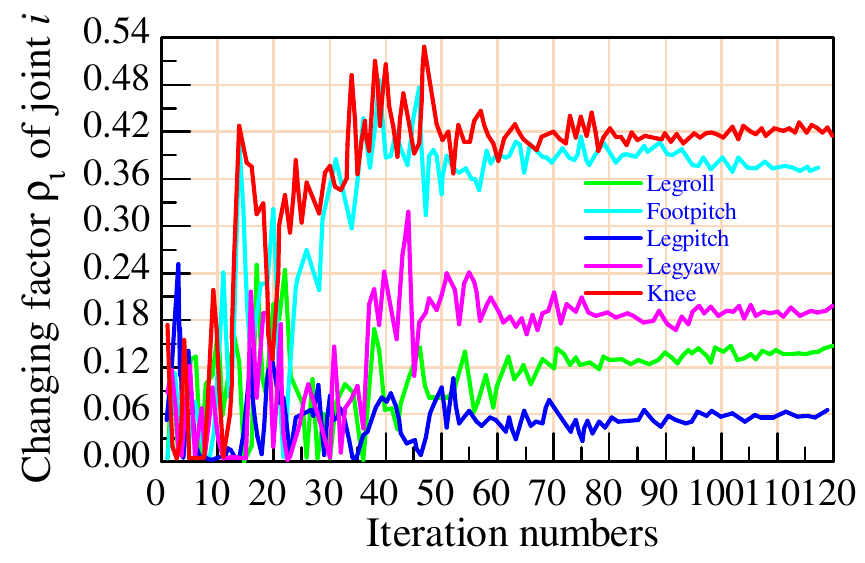}}               %
\caption[]{\small ({\bf A}) Gait trajectory curves of joint rotation angles at the right leg of humanoid robotics Nao
and relationship between the optimized joint angles of walking humanoid robotics against time. This example was
constructed under the following conditions: 1) Hip-joint height $h$ of humanoid robotics from the ground $h=518[mm]$,
2) legged walking step length $s=100[mm]$, 3) Duty-ratio $d=0:7$. ({\bf B}) Adaptive evolution of the different joints
with the best gait for walking over time. Obviously, the most human-like walking and stable gait are optimized,
especially ${\color{blue}\bf knee~joint}$ can be highly adapted.}%
\label{efig03}
\end{figure*}

Before optimizing physical robotics using the EACO, we select the different parameters of the EACO algorithm
and test the physical humanoid robot {\color{blue}NAO} with a height of 57cm, a mass of 4.5kg and 22 degrees of freedom
(http://www.aldebaran-robotics.com), and then the optimized results of gait and motion trajectory of physical humanoid robot {\color{blue}NAO}
are shown in Fig.~{{\ref{efig02}}}.~Gait trajectory curves of rotation angles at right leg joint of humanoid robotics Nao and relationship between the optimized joint angles of walking humanoid robotics against time are illustrated in Fig.~{{\ref{efig03}}}({\bf A}) and figs.~{\color{red}{\textrm S5}}({\bf A}) and fig.~{\color{red}{\textrm S5}}({\bf B}), and~the adaptive evolution results of the different joints with the best walking gait over time are shown in Fig.~{\color{red}{\ref{efig03}}}({\bf B}). Obviously, the most human-like and stable gait for legged walking are optimized, especially {\color{blue}{knee joint}} can be highly adapted~{\cite{gong01,farzadpour01}}.

\begin{figure*}[ht]
\centering
\subfloat[]{\includegraphics[width=5.1cm,height=3.2cm]{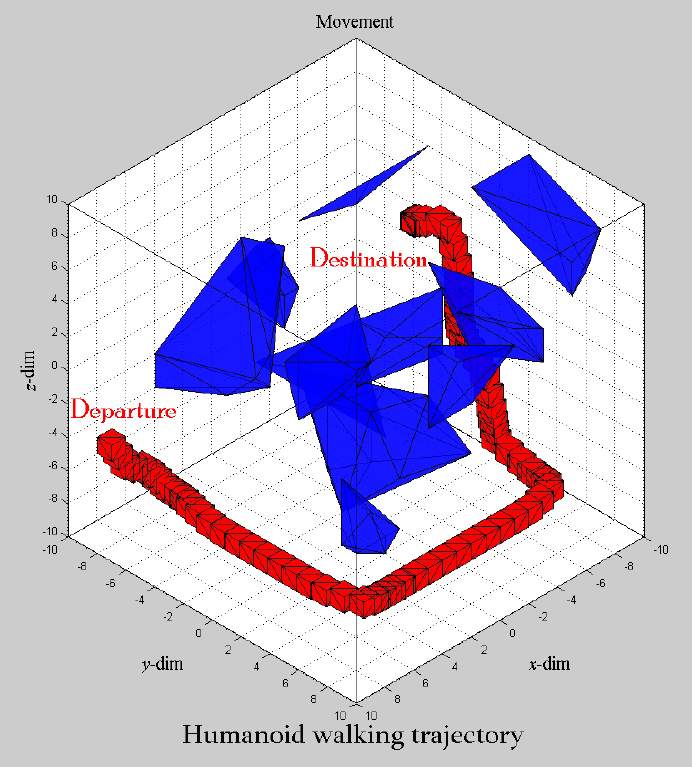}} \hspace{1pt}%
\subfloat[]{\includegraphics[width=5.1cm,height=3.2cm]{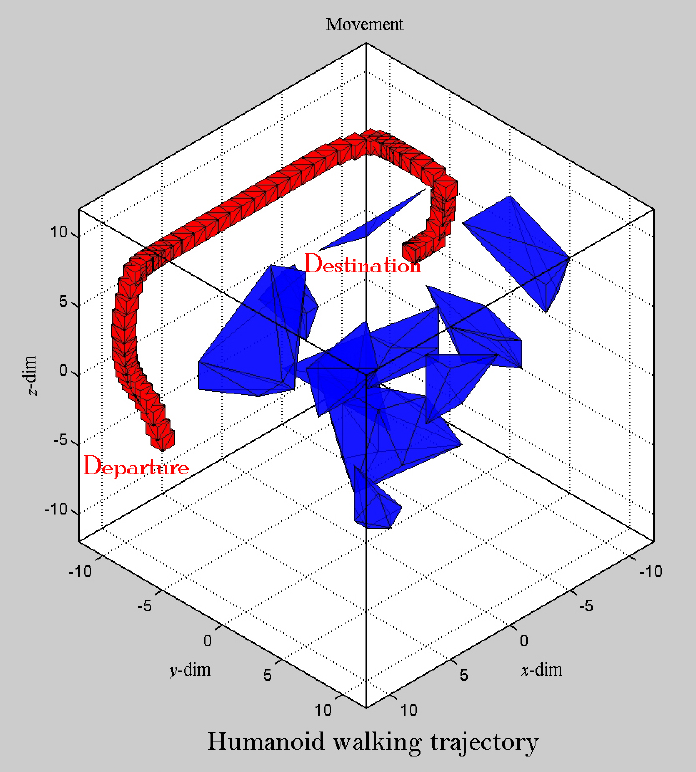}} \hspace{1pt}%
\subfloat[]{\includegraphics[width=6.0cm,height=3.2cm]{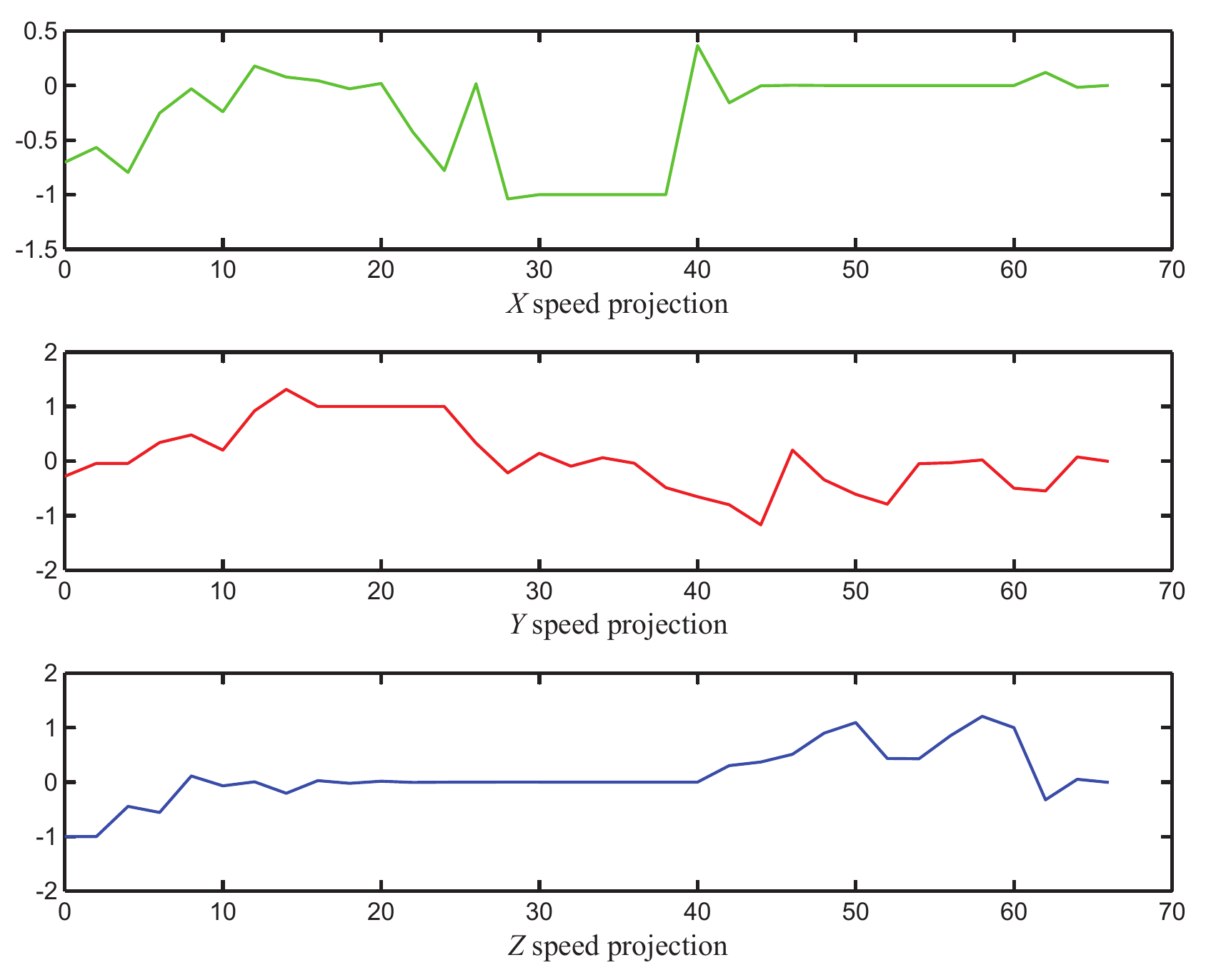}}
\vspace*{0.2cm}
\caption[]{{\bf These simulated experiments}. These experiments were done in the complex environments with
a different number of three-dimensional polygon obstacles, different population sizes, and a different number of iterations using only genetic algorithm and the EACO with genetic and crossover operators respectively for legged walking trajectory planning and gait parameters’ optimization of the simulated humanoid robot.
({\bf A}) Walking trajectory of the simulated humanoid robotics using only genetic algorithm is not
optimal. ({\bf B}) Obviously, the walking trajectory of the simulated humanoid robotics in the same
environments using the EACO with the mutation and crossover operators is optimal or suboptimal. ({\bf C}) speed projects of
the simulated humanoid robot walking in $x$, $y$, and $z$ directions by using the EACO with the mutation and crossover operators.}
\label{efig05}
\end{figure*}

\begin{figure*}[ht]
\vspace*{-0.2cm}
\centering
\subfloat[frame368]{\includegraphics[width=5.4cm,height=3.2cm]{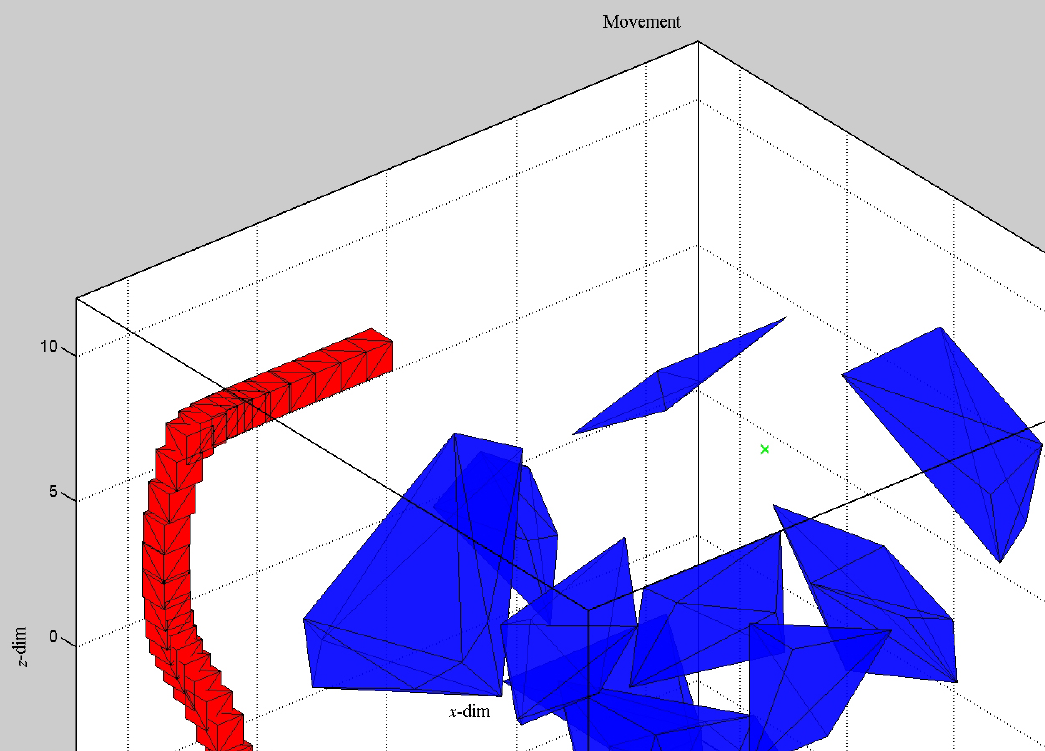}}  \hspace{0pt}
\subfloat[frame986]{\includegraphics[width=5.4cm,height=3.2cm]{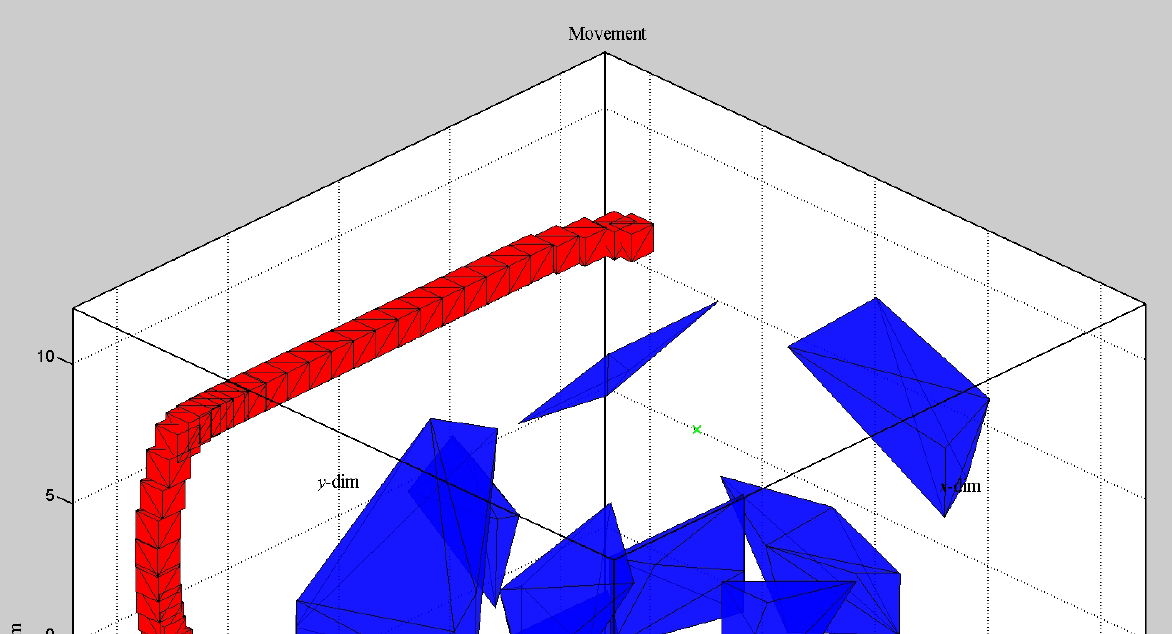}}  \hspace{0pt}
\subfloat[frame1872]{\includegraphics[width=5.4cm,height=3.2cm]{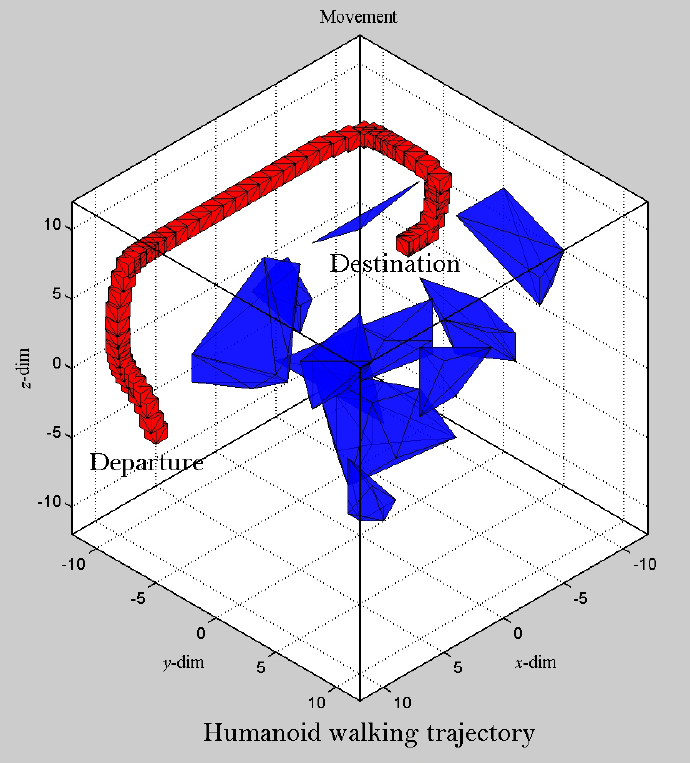}}
\vspace*{0.1in}
\caption[]{{\bf The experiments on a simulated humanoid robot in the complex environments (VIDEO{\color{red}~\bf{I}}).}
Based on genetic algorithm, the experiments were realized in the complex environments with a different number of three-dimensional polygon obstacles, different population sizes, and a different number of iterations for the walking trajectory and
gait parameters’ optimization of the simulated humanoid robot. Obviously, the walking trajectory of the simulated
humanoid robotics using only genetic algorithm is not optimal or suboptimal in the same environments.}
\label{efig06}
\end{figure*}

\begin{figure*}[ht]
\centering
\vspace*{-0.4cm}
\subfloat[frame198]{\includegraphics[width=5.4cm,height=3.2cm]{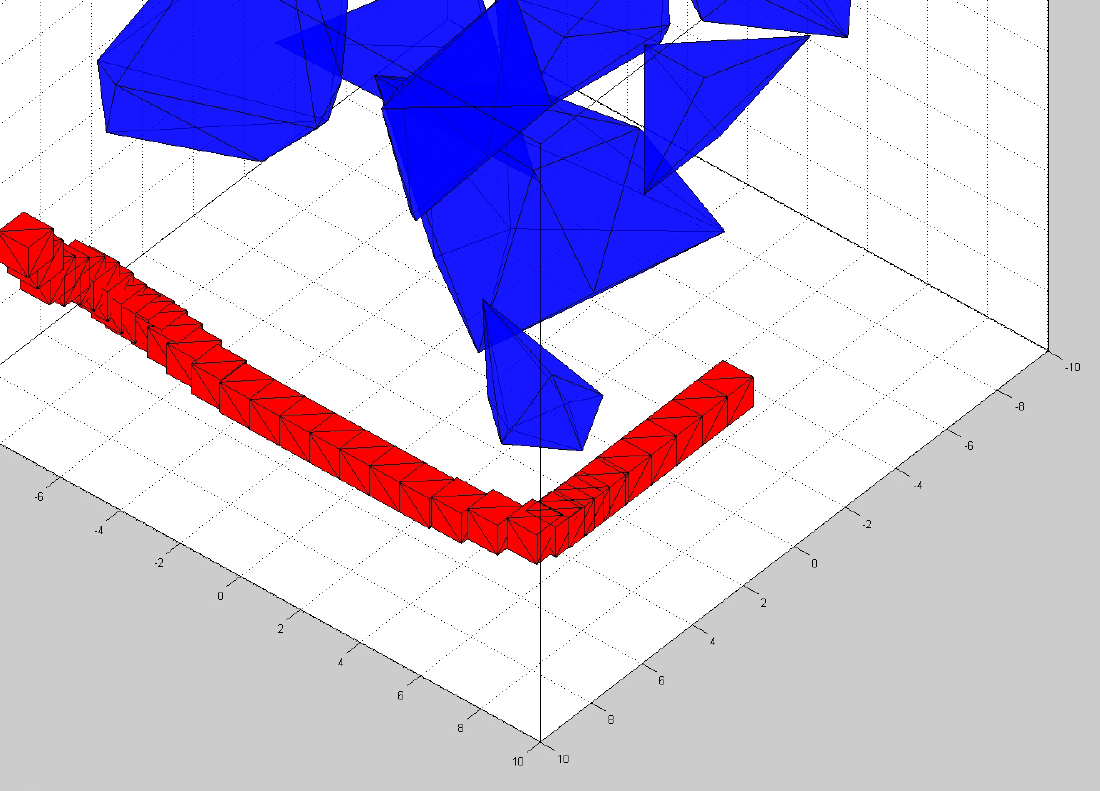}} \hspace{0pt}
\subfloat[frame827]{\includegraphics[width=5.4cm,height=3.2cm]{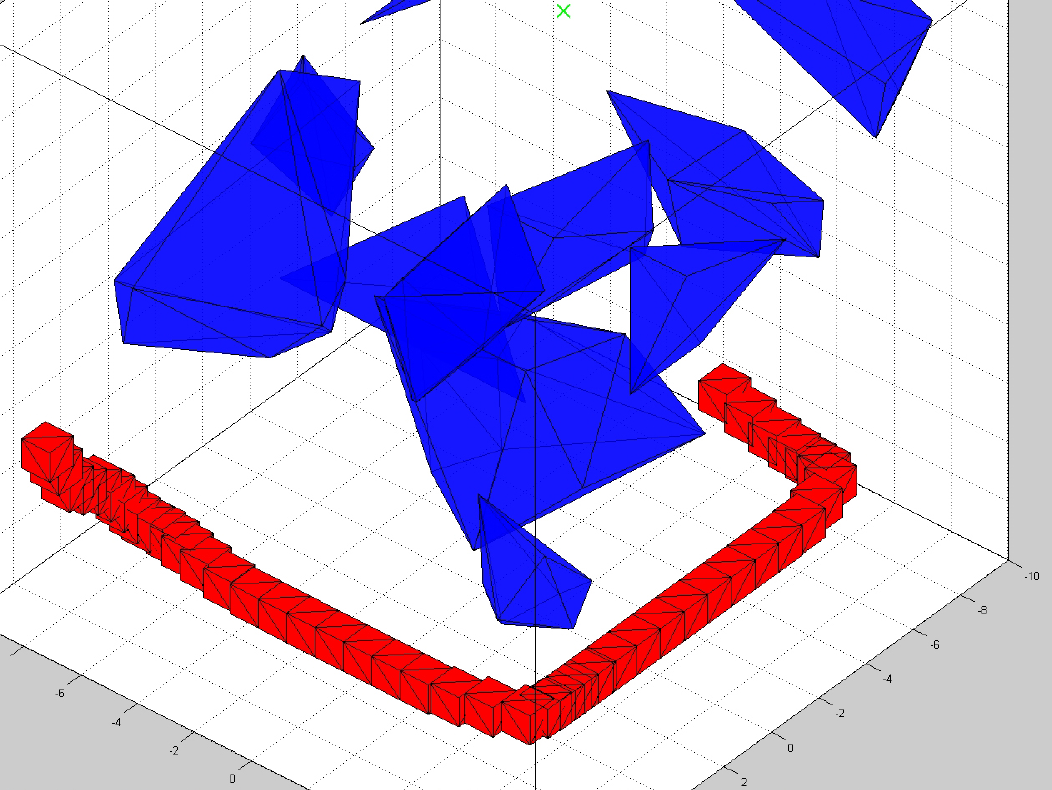}}   \hspace{0pt}
\subfloat[frame1488]{\includegraphics[width=5.4cm,height=3.2cm]{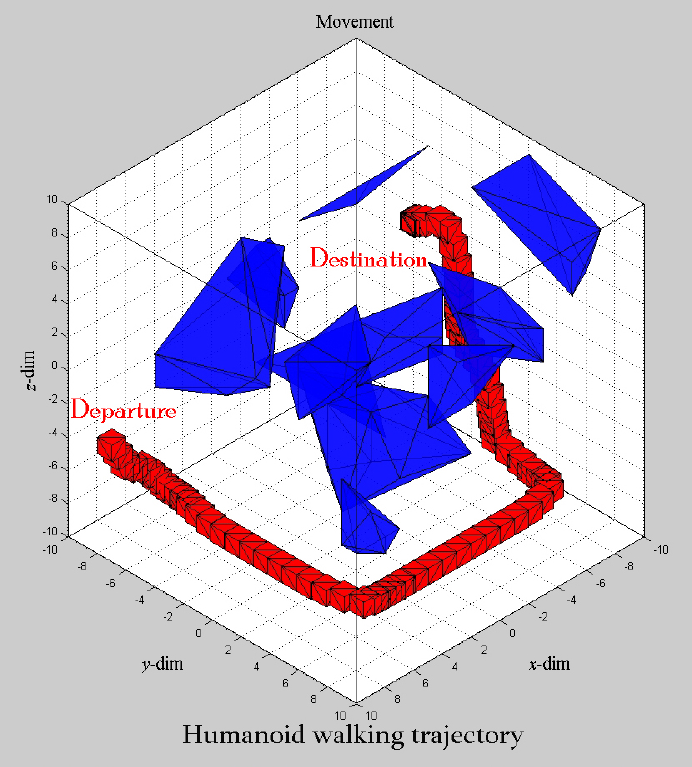}}
\vspace*{0.1in}
\caption[]{{\bf\small{The experiments on a simulated humanoid robot in the complex environments (VIDEO {\color{red}\bf{II}})}}.
Based on the EACO, the experiments were done in the complex environments with a different number of three-dimensional polygon obstacles,
different population sizes, and a different number of iterations with genetic and crossover operators for legged walking trajectory planning and
gait parameters’ optimization of the simulated humanoid robot. Obviously, the walking trajectory of the simulated
humanoid robotics is optimal or suboptimal using the EACO with the mutation and crossover operators.}
\label{efig07}
\end{figure*}

\begin{figure*}[ht]
\vspace*{-0.2cm}
\centering
\subfloat[frame2688]{\includegraphics[width=5.4cm,height=3.2cm]{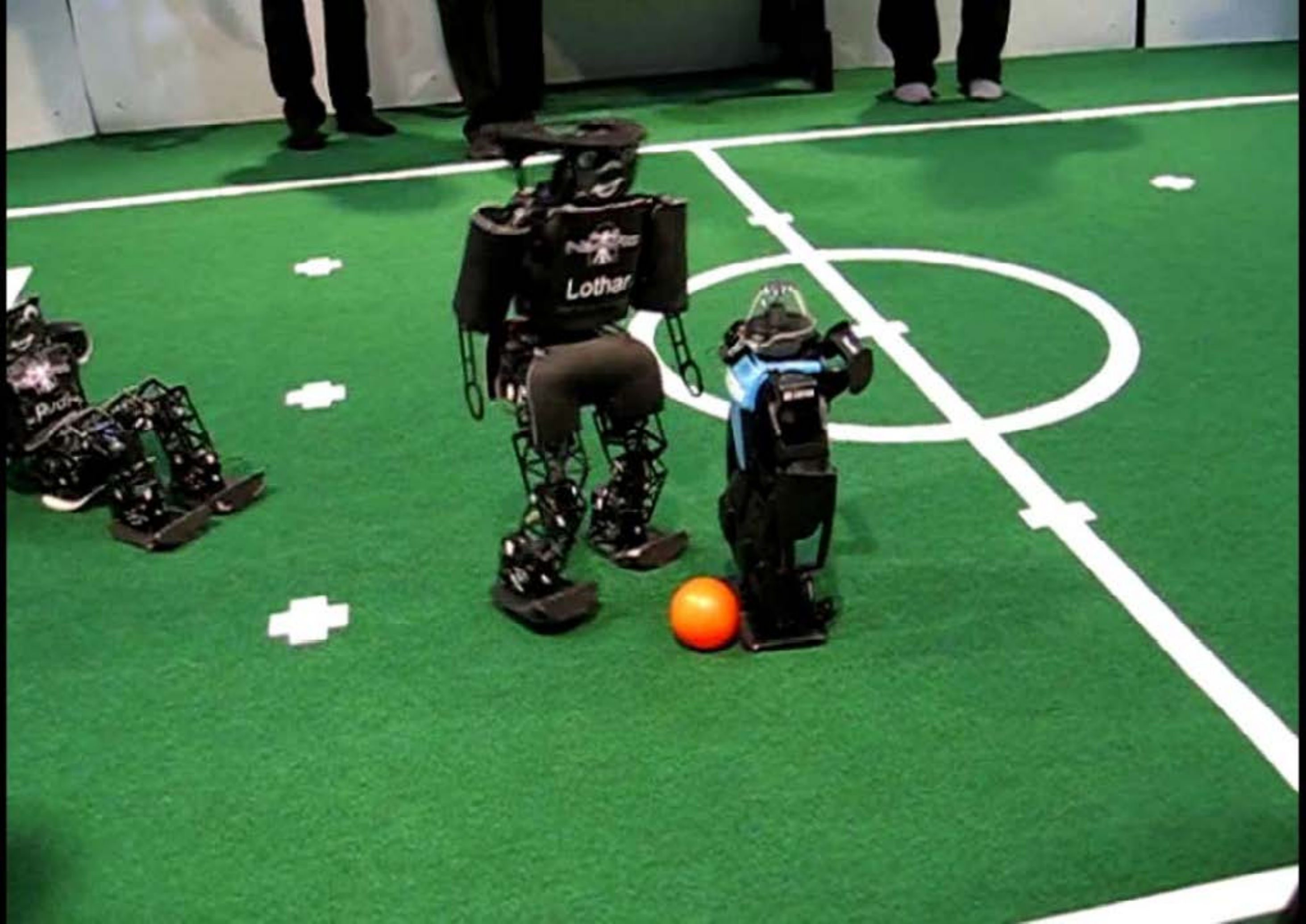}}  \hspace{0pt}
\subfloat[frame3272]{\includegraphics[width=5.4cm,height=3.2cm]{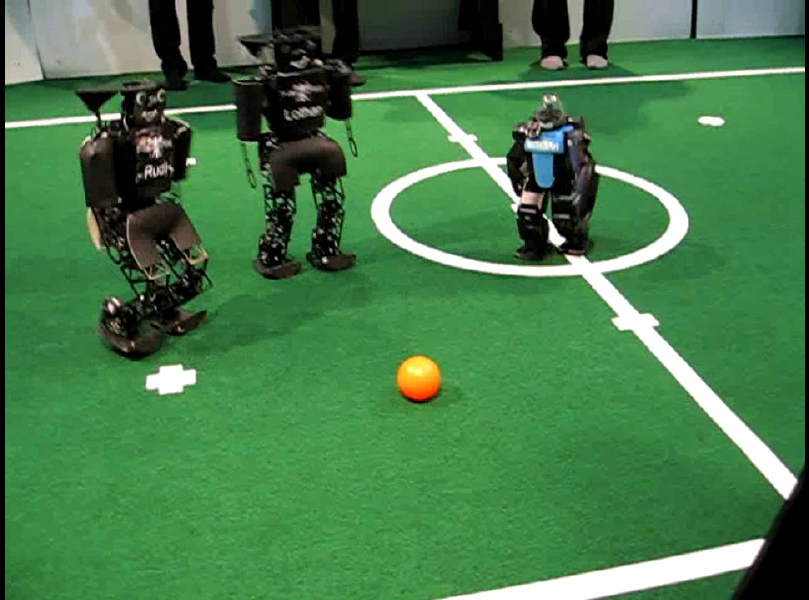}}  \hspace{0pt}
\subfloat[frame3898]{\includegraphics[width=5.4cm,height=3.2cm]{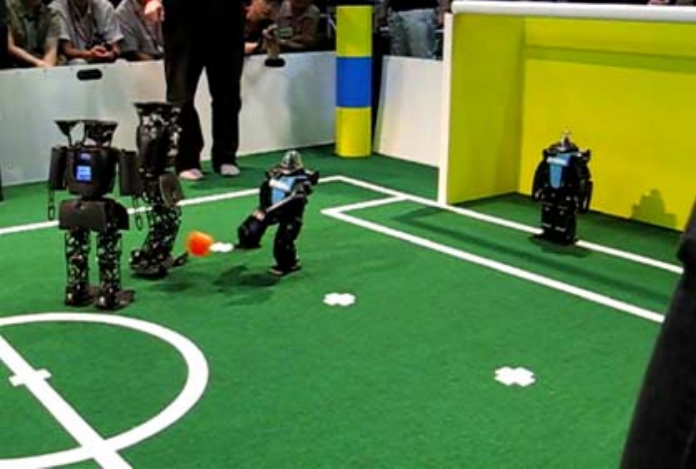}} \vspace{0pt} \\
\subfloat[frame9112]{\includegraphics[width=5.4cm,height=3.2cm]{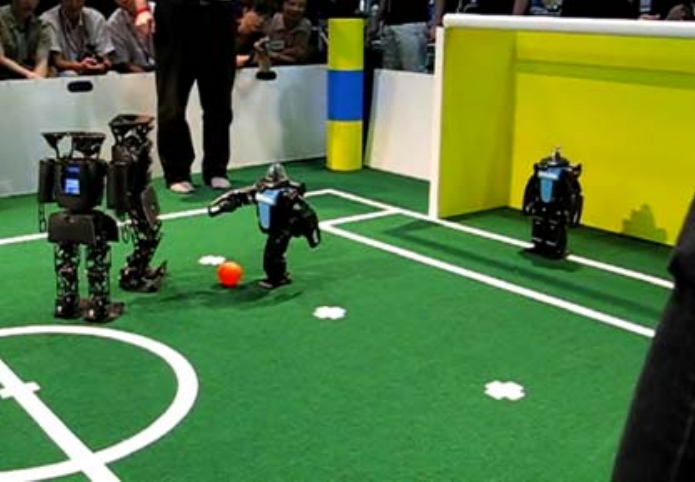}} \hspace{0pt}
\subfloat[frame9224]{\includegraphics[width=5.4cm,height=3.2cm]{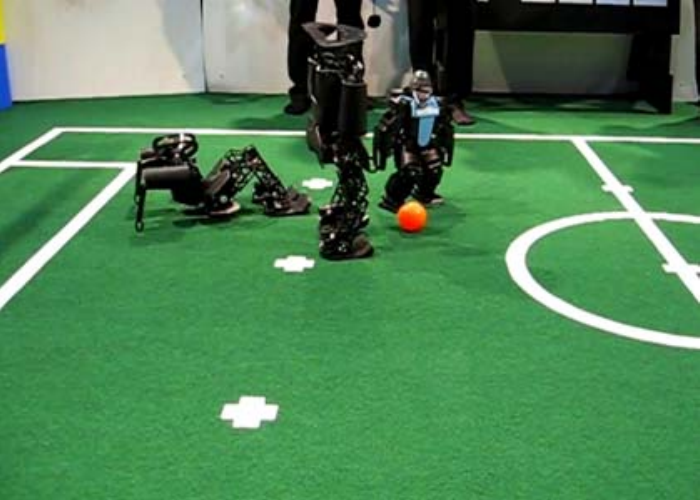}} \hspace{0pt}
\subfloat[frame9348]{\includegraphics[width=5.4cm,height=3.2cm]{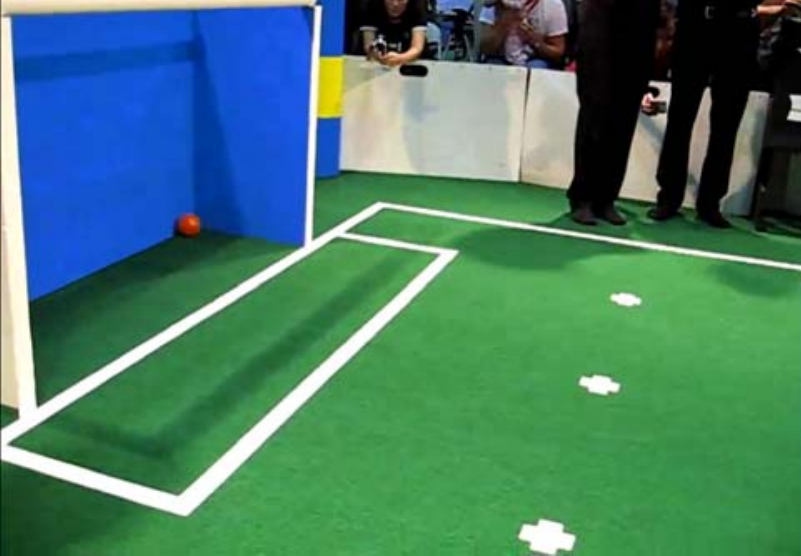}} \hspace{0pt}
\vspace*{0.1in}
\caption{{\bf The real humanoid soccer robotics competition (VIDEO {\color{red}\bf{III}})}. Physical humanoid robotic soccer competition images abstracted from Video {\color{red}\bf{III}} as shown in (\bf A)--(\bf F)}.
\label{efig08}
\end{figure*}

Figs.~{\color{red}{\ref{efig04}}}({\bf A}) and {\color{red}{\ref{efig04}}}({\bf B}) illustrate respectively
how the performance for solving optimization problems varies as the crossover rate and the mutation rate increase. Fig.~{\color{red}{\ref{efig05}}}
shows the optimized walking trajectory of the simulated humanoid robot in the complex environment with ten three-dimensional polygon obstacles that we designed and implemented, with ({\bf{A}}) walking trajectory based on the genetic algorithm; ({\bf{B}}) the optimized walking trajectory using the EACO with mutation and crossover operators; ({\bf{C}}) The speed projects of the simulated humanoid {\color{blue}NAO} walking in $x$, $y$, and $z$ directions. The images extracted from VIDEO of the walking trajectory of the simulated humanoid robot using only genetic algorithm demonstrated that the walking trajectory is not optimal or suboptimal (Fig.~{\color{red}{\ref{efig06}}}) and the walking trajectory of the simulated humanoid robot using the EACO with genetic and crossover operators is optimal or suboptimal (Fig.~{\color{red}{\ref{efig07}}}). The real experiments on physical humanoid robotic soccer competition are illustrated in Figs. {\color{red}{\ref{efig08}}}({\bf A})--{\color{red}{\ref{efig08}}}({\bf F}) extracted from VIDEO {\color{red}\bf{III}}.

The simulated experiments that we designed and implemented using~{\color{blue}MATLAB 2016B} and~{\color{blue}WEBOTS Pro 8.6.1} simulator which includes the
physical models based on rigid body dynamics can simulate varieties of robots defined by users. The goal is to find out
how the results differ when the sizes of ant colony and the forms of the neighborhood within the ant colony varied. In order to
determine whether these parameters are the optimized walking parameters within this range, our tests set maximum number of iterations
to 40 and 80 respectively. Lately similar tests on the humanoid robots choose an accelerated walking with a limited time,
and the fitness value is defined by the distance walking in the desired walk direction during the time limit of 20 seconds.
All parameters of the EACO take the values recommended by~{\color{blue}\cite{niehaus02}}.

\section{CONCLUSIONS AND DISCUSSION}
\begin{figure*}[!t]
\centering
\subfloat[]{\includegraphics[width=8.1cm,height=4.0cm]{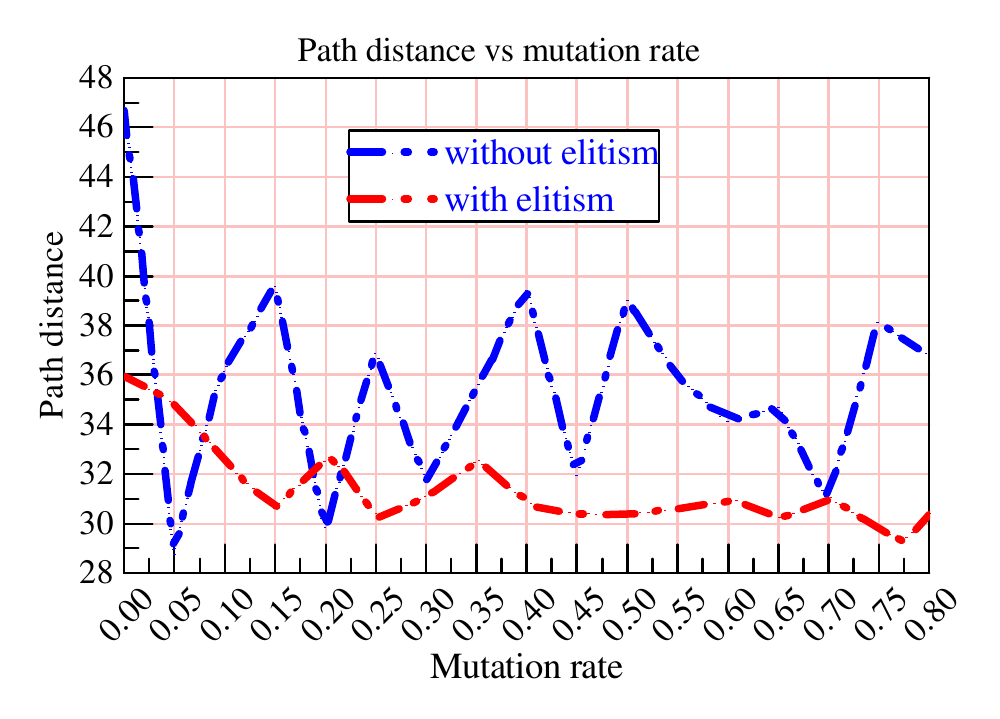}}  \hspace{1pt}
\subfloat[]{\includegraphics[width=8.1cm,height=4.0cm]{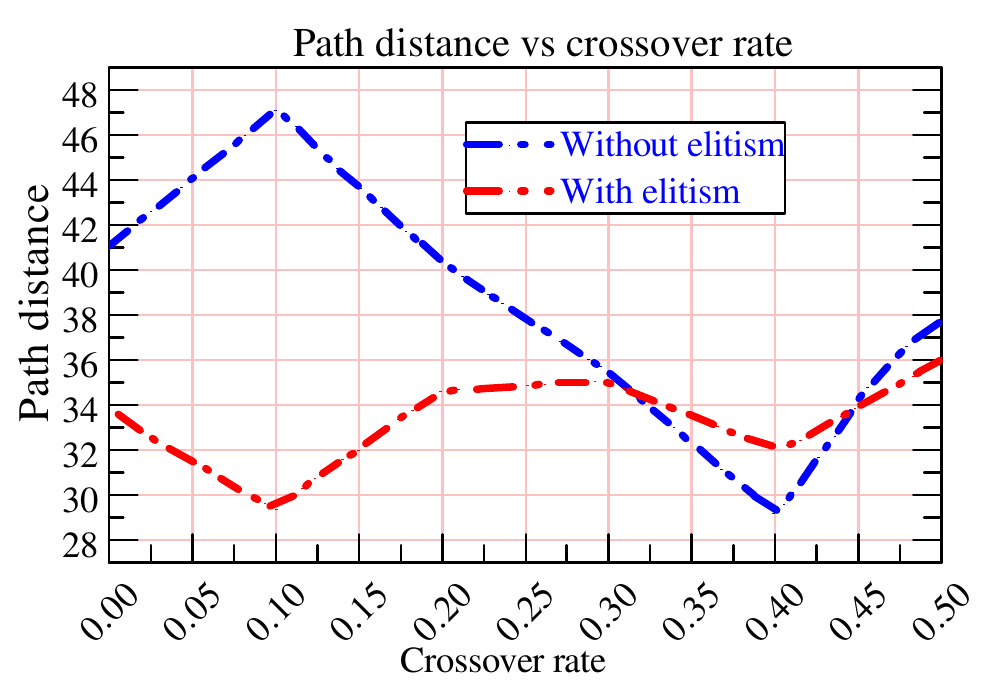}}  \hspace{1pt}
%
\caption{{\bf Path distance closely related to crossover rate and mutation rate.} In complex environments where the size and number of obstacles are the same as the simulated experiments (Video {\color{red}\bf I} and Video {\color{red}\bf II}). ({\bf A}) illustrates how the performance (path distance from starting point to destination) varies as the crossover rate increases. ({\bf B}) shows how the performance varies as the mutation rate increases.}
\label{efig04}
\end{figure*}

\begin{figure*}[ht]
\vspace*{-0.2in}
\centering
\subfloat[]{\includegraphics[width=8.1cm,height=4.0cm]{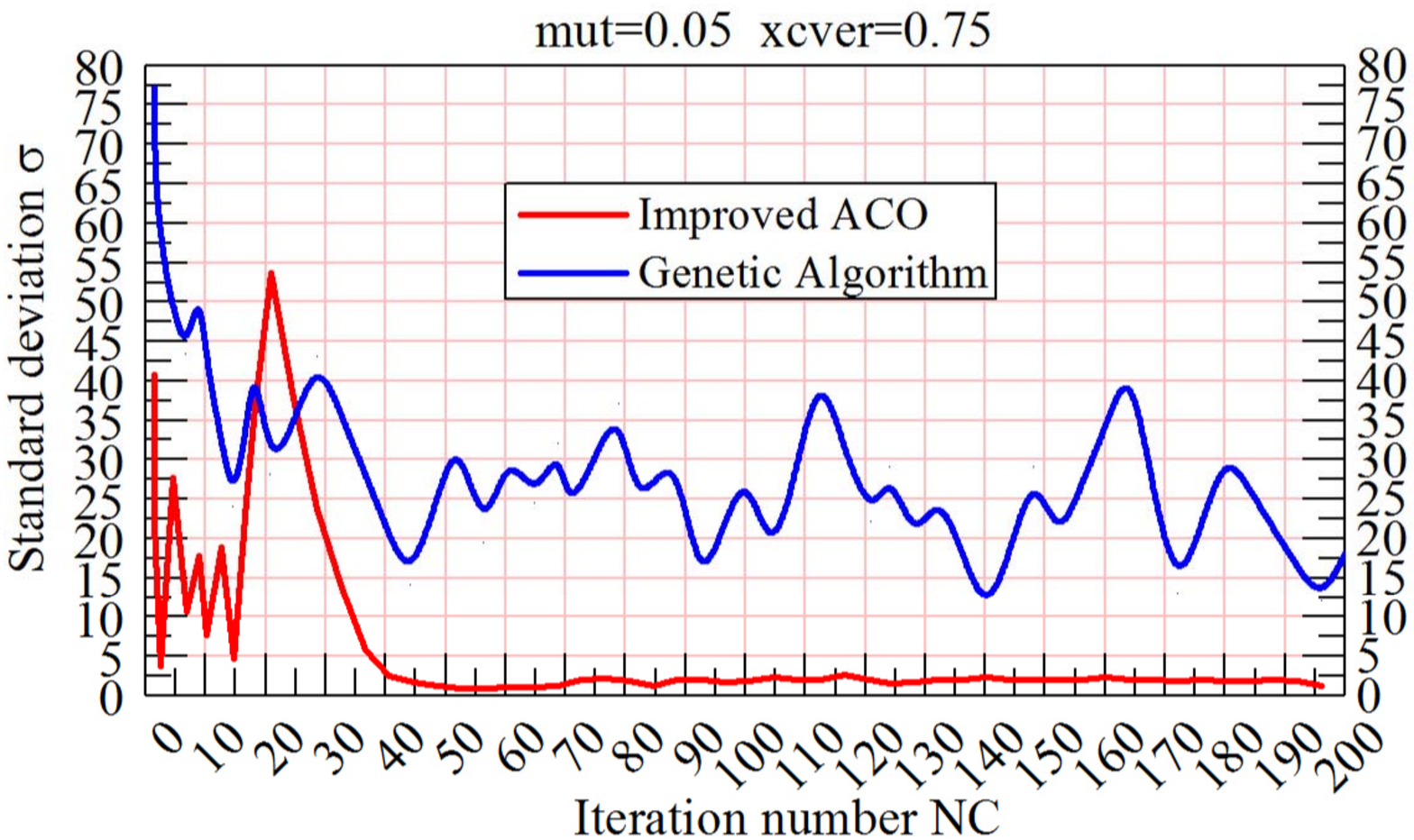}} \hspace{1pt}%
\subfloat[]{\includegraphics[width=8.1cm,height=4.0cm]{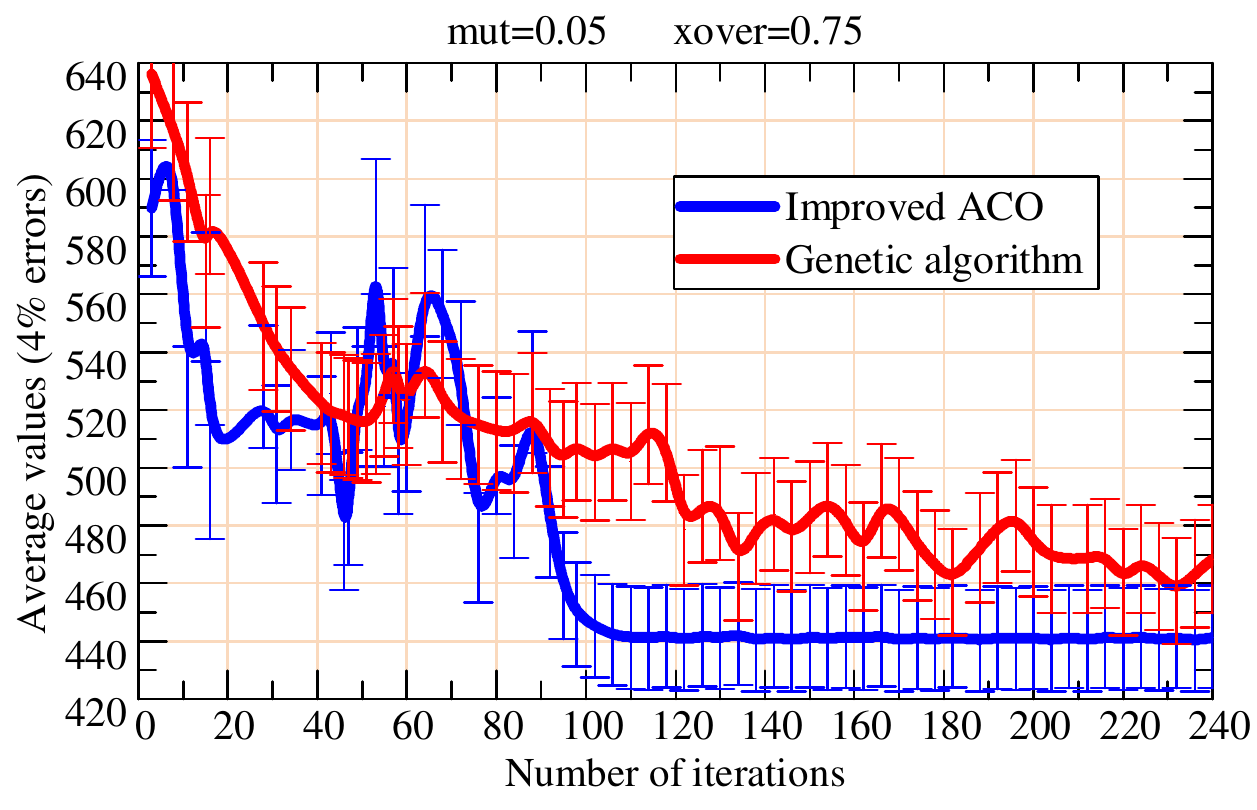}}
\vspace*{-0.1in}
\caption[]{\small{Effect of the standard deviation ${\color{blue}{\sigma}}$ on convergence rates of the EACO and the GA.} (A) the ${\color{blue}{\sigma}}$ values of the EACO tended towards stability when the number of iterations NC$\geq$45. (B)
The average values of the EACO tended to stabilize while the number of iterations$\geq$108.}
\label{efig09}
\end{figure*}

Here, we have demonstrated the applicability and the effectiveness of the EACO algorithm applied to robotics control optimization.
As a matter of fact, the EACO algorithm can reduce the premature convergence probability and promote the convergence rate of the EACO system
by preserving the population diversity, namely, the dynamic application of mutation and crossover operators (Figs.~{\color{red}{\ref{efig01}}}({\bf A}) and ~{\color{red}{\ref{efig01}}}({\bf B})) and an elitist strategy (Figs.~{\color{red}{\ref{efig04}}}({\bf A}) and {\color{red}{\ref{efig04}}}({\bf B})).
Therefore, the algorithms prevented effectively the systems from getting trapped into local optima and achieved optimal
solutions or suboptimal solutions in most cases. Furthermore, we can obtain the global optimal solutions by self-adjusting the path
searching behavior of robotics in accordance with the objective function. In this work, an elitist
strategy is used in the EACO algorithm to enhance its convergence performance, as shown in
Fig.~{\color{red}{\ref{efig04}}} and Fig.~{\color{red}{\ref{efig09}}}. As can be seen from Fig.~{\color{red}{\ref{efig04}}}
and Fig.~{\color{red}{\ref{efig09}}}, two major factors of affecting
convergence speed of the EACO: ($i$) the pheromone density deposited on the paths and ($ii$) heuristic
functions. Selecting appropriate heuristic function is essential for promoting the convergence rate.
If the heuristic function value is too large, it will inhibit the effects of pheromone, and the heuristic function value
is too small, it affects the convergence rate. Moreover, the dynamic nearest-neighbor selection strategy can be used to promote the convergence speed.

To enhance global searching capability and promote convergence speed, the EACO algorithms are performed
by adaptively updating pheromone trails using adaptive global update rules. First, we introduce the mutation
operator which is advantageous to the ACO algorithms and can enhance capability of the EACO algorithms to preclude
local optima and promote convergence rate by assigning a processor to each colony. Then our algorithm is able
to prevent excessively concentrated searching by enhancing global searching capability of the ACS at the initial
stage and promote the convergence rate in the later stages using changeable volatility coefficient.
Finally, the EACO approach is used to optimize walking stability, walking velocity, and energy consumption
of humanoid robotics {\color{blue}\cite{zhao01}}. The optimized parameters are applied to the normal walking experiments
for the simulated and real robots. The locomotion performance with optimized parameters is superior to that with non-optimized parameters.
The EACO algorithm on a wide range of problems, from ACO, real-coded GA, simulated annealing to complex robotics systems is validated.
The experimental results illustrate the capability for this method to discover the premature convergence probability, tackle successfully inherent stagnation and poor convergence rate of ant colony optimization, and promote the convergence rate of the EACO-based humanoid robotics systems.
The results tested on combinatorial optimization problems have demonstrated that global best solution and the
convergence rate have been promoted. One of the main advantages of the EACO is to confirm the possibility to obtain the optimized results.

\begin{figure*}[!ht]
\vspace*{-0.1in}
\centering
\subfloat[]{\includegraphics[width=8.1cm,height=4.0cm]{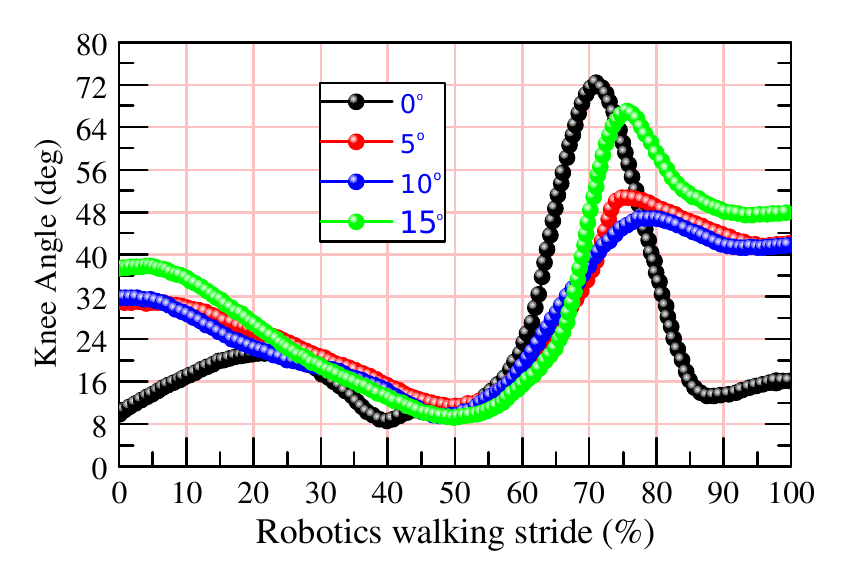}}    \hspace{1pt}    
\subfloat[]{\includegraphics[width=8.1cm,height=4.0cm]{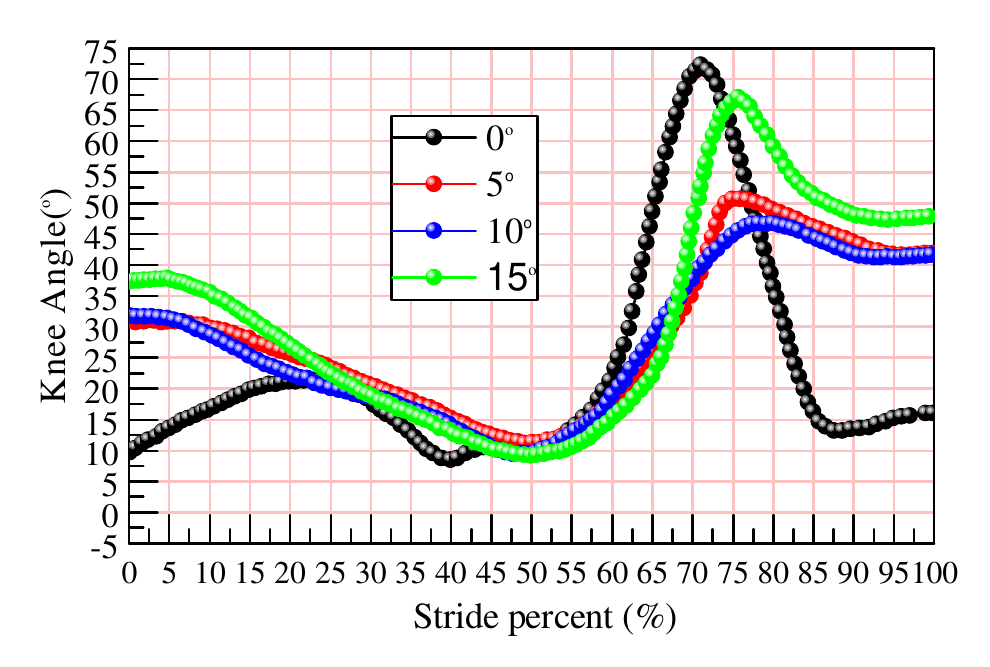}}    \hspace{1pt} \\ 
\subfloat[]{\includegraphics[width=8.1cm,height=4.0cm]{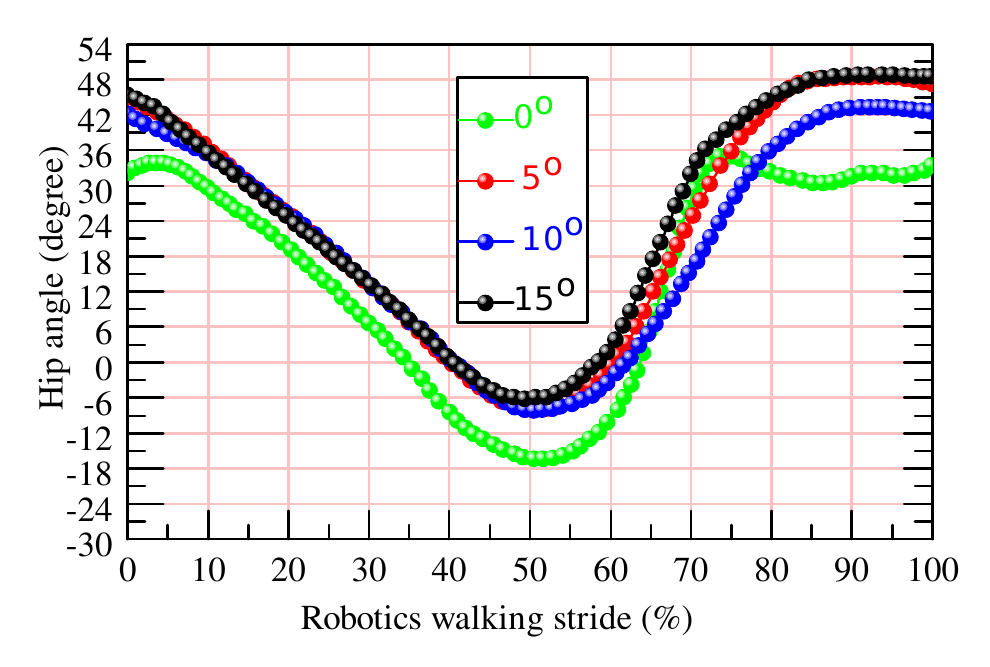}}    \hspace{1pt}    
\subfloat[]{\includegraphics[width=8.1cm,height=4.0cm]{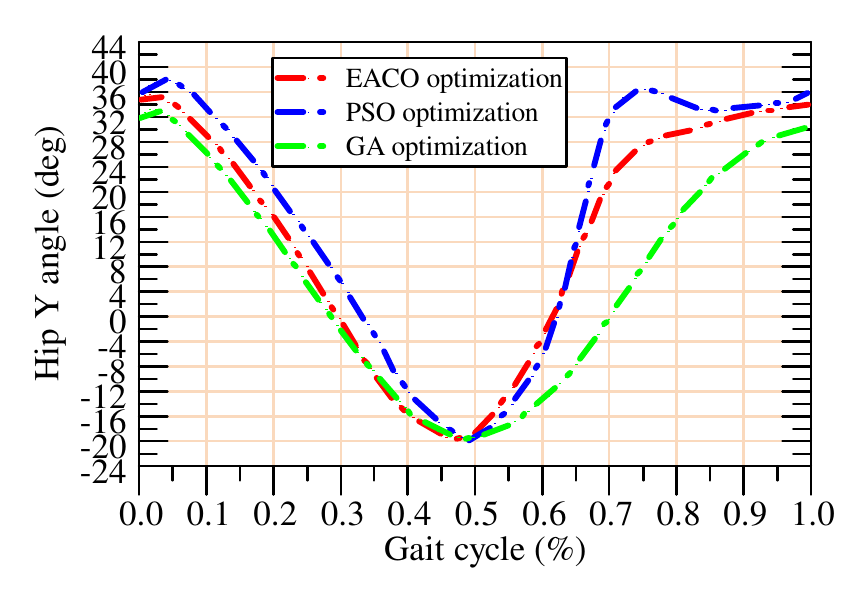}}    \hspace{1pt} \\
\subfloat[]{\includegraphics[width=8.1cm,height=4.0cm]{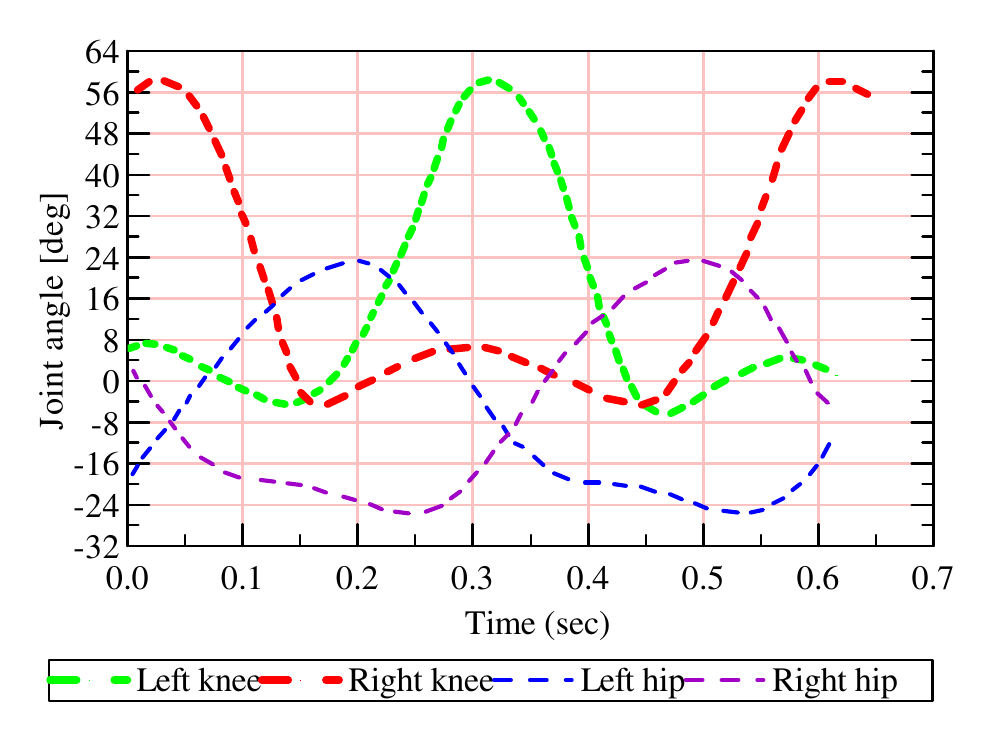}}    \hspace{1pt}
\subfloat[]{\includegraphics[width=8.1cm,height=4.0cm]{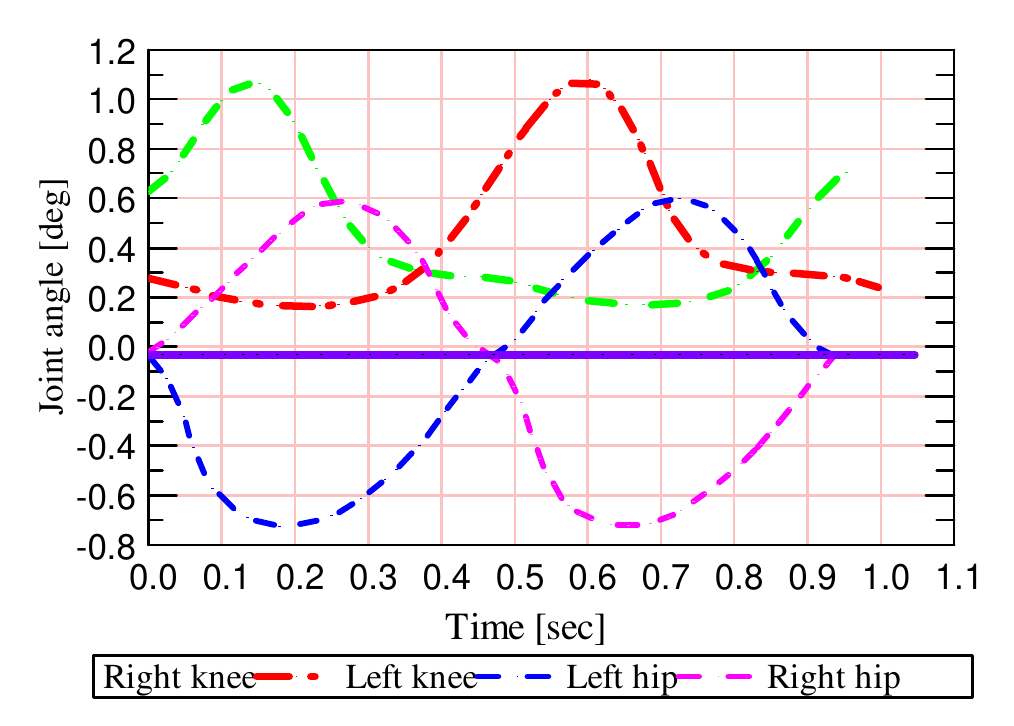}}    \hspace{0pt}
\vspace*{-0.1in}
\caption[]{{\bf\small{Bipedal locomotion controll.}} {\small{The angular trajectories of ankle, knee, and hip of humanoid robotics walking on the different slope roads are illustrated in ({A}), ({B}), and ({C}). The optimized trajectories of hip angles with the EACO, the PSO, and the GA are shown in (D). The walking angular trajectories of knee and hip captured from human walking are presented in (E), The angles of knee and hip joints in one period of human walking signals are shown in (F).}}}
\label{efig10}
\end{figure*}

Nevertheless, we argue that the EACO algorithm for solving combinatorial optimization problems has the potential to be more widely used in practice than other approaches because it have several advantages: (i) Three kinds of optimization algorithms converge to optimal solution or sub-optimal solution. (ii) The enhanced EACO algorithm is recognized to be an efficient method for solving combinatorial optimization issues, as compared with other three methods, although the iteration number required by this approach is only slightly less than that
required by {\color{blue}Tabu} search. (iii) Obviously, our algorithm has higher computational complexity as compared with the other two algorithms. Moreover, its parameters should be tuned carefully and optimized. (iv) We developed the EACO-based optimization approach to walking stability, walking velocity, and energy consumption of humanoid robotics. The optimized parameters are applied to the normal bipedal walking experiments on the simulated and real robots. The angular trajectories of {\color{blue}\bf knee}, {\color{blue}\bf ankle}, and {\color{blue}\bf hip} of humanoid walking on the different slope roads are illustrated respectively in Figs.~{\color{red}{\ref{efig10}}}({\bf A}), {\color{red}{\ref{efig10}}}({\bf B}), and {\color{red}{\ref{efig10}}}({\bf C}). Comparison of the trajectories of {\color{blue}\bf hip} angles in bipedal walking with the EACO,
the PSO~{\color{blue}\cite{ghashochibargh01}}, and the GA optimizations is shown in Fig.~{\color{red}{\ref{efig10}}}({\bf D}). The walking angular trajectories of {\color{blue}\bf knee} and {\color{blue}\bf hip} captured from human walking are presented in Fig.~{\color{red}{\ref{efig10}}}({\bf E}). The angles of knee and hip joints in one period of human walking signals are shown in Fig.~{\color{red}{\ref{efig10}}}({\bf F}). The locomotion performance with optimized parameters is superior to that with non-optimized parameters.

We will submit our original papers in the future which can truly achieve the combination of basic science and robotics to develop suitable fabrication and assembly strategies {\color{blue}\cite{yang10}}, to tackle questions of control, navigation, optimization, and communication.


\section{MATERIALS AND METHODS}
\subsection{Algorithm Modeling}
~The combinatorial optimization model $({\bf{S}},\Omega,f)$ is mapped to the following characteristic problems:
\begin{itemize}
\item {A discrete set $\xi$=$\{c_1,c_2,\cdots,c_{N}\}$ represents the composition of an optimization problem;}
\item {Based on all possible sequences $x$=${\langle}c_i$, $c_j,\cdots,c_k,\cdots{\rangle}$ we define
the finite state set $\chi$ of the question, where ${\langle}c_i,$$c_j,\cdots,c_k,\cdots{\rangle}$ is element of $\xi$.
The sequence length is defined as $|x|$ which denotes the number of the elements in the sequence, and its maximum value $l<\infty$;}
\item {The candidate solution set ${S}$ is a subset of the finite state set $\chi$, namely, $S{\subseteq}\chi$;}
\item {The feasible state sets $\widetilde{\chi}{\subseteq}{\chi}$ is constructed by sequence $x{\in}\widetilde{\chi}$
which can satisfy the constraint condition $\Omega$;}
\item {If $S^{\ast}{\subseteq}\widetilde{\chi}$ is satisfied and $S^{\ast}{\subseteq}S$, then nonempty set $S^{\ast}$
is regarded as an optimal solution.}
\end{itemize}
According to the description above, the combinatorial optimization problems
may be denoted by the oriented graph $\bf{G}$$=$$({\bf \zeta},{\bf L},{\bf P})$, $\bf\zeta$ indicates the node set,
$\bf L$ represents the linked arcs set between all nodes, ${\bf P}$ is the pheromone vector set.

In ant colony optimization, each ant searches and chooses the initial node in the graph $\bf G$,
stochastically passes the linked arcs and chooses the next node
according to the pheromone intensity on the linked arcs.
where $\Omega$ may guarantee the node in the feasible solution set. Once the ant
completes one cycle search, then the pheromone is updated.
\subsection{Design and Implementation of EACO}
To facilitate the understanding and explanation of the proposed algorithm, the description and implementation of the EACO algorithm at each generation is summarized into the following steps:
\begin{itemize}
\item[1)]{\bf Initialization:}
Create initial feasible solutions and determine the corresponding objective function value. Set up
initial values of pheromone and other variables.
%
\item[2)]{\bf Local update:}
Motivation of the local pheromone update is to preclude stagnation, reduce the pheromone density of the visited edges, and make
those edges attract no longer other ants. The local pheromone update procedure is activated by sending some local ants to the neighborhood
of the selected global ants. When each ant uses the edge $e_{ij}$ in its tour, its pheromone trails $\tau_{ij}$ are updated as:
\begin{equation}
\tau_{ij}(t+1)=(1-\rho)\tau_{ij}(t)+\rho\Delta\tau_{ij}.
\label{eqn:68}
\end{equation}
\begin{equation}
\Delta\tau_{ij}=\sum\limits^m_{k=1}\Delta\tau_{ij}^k.
\label{eqn:69}
\end{equation}
\setlength{\arraycolsep}{0.0em}
\begin{eqnarray}
\Delta\tau_{ij}^{(k)}=\bf\left\{\begin{array}{ll}
\frac{Q}{L_k} & \textrm{~if the ant $k$ uses~}~\textrm{edge $e_{ij}$ in its tour} \\
~0            & \textrm{~otherwise}
    \end{array} \right.
\label{eqn:70}
\end{eqnarray}
\end{itemize}
where:
\begin{itemize}
\item $\rho{\in}(0,1)$ represents the coefficient which is called local pheromone persistence parameter, thus (1-$\rho$) corresponds to evaporation
ratio of the pheromone trails from $t$ to $t+1$. It turns out that reduction of $\rho$ values favors convergence of the EACO algorithms
to the best solution. Obviously, as $\rho$ value decrease, the pheromone trail changes at each iteration reduce, such that to get
substantial change level of the pheromone trails intensity a larger number of iteration computations
must be executed. Briefly, for $\rho\rightarrow1$ a small amount of pheromone is decayed, the
convergence rate gets slower. Conversely for $\rho\rightarrow0$, more pheromone is decayed leading
to faster convergence~{\color{blue}\cite{zecchin01}}.
\item $Q$ is a factor of local pheromone update which may affect the convergence rate of the ACO algorithm.
\item $L_k$ denotes path length ant $k$ traveled at current iteration.
Once all ants complete their iterations, global pheromone updating occurs, this is computed by~Eq.~[{\color{red}\ref{eqn:07}}].
\end{itemize}
\begin{itemize}
\item[3)]{\bf Global update:}    
When all ants complete their tours of all nodes, the global pheromone update can be performed
based on the global pheromone update rules. Thus, the pheromone trail intensity is updated by the following equation:
\begin{equation}
\tau_{ij}(t+1)=(1-\alpha)\tau_{ij}(t)+\alpha\Delta\tau_{ij}{(t)}.
\label{eqn:71}
\end{equation}
\begin{displaymath}
\Delta\tau_{ij}=\left\{\begin{array}{ll}
\frac{1}{L_{gb}} & \textrm{if~an edge $e_{ij}$ $\in$ global best tour}          \\
~0               &\textrm{otherwise}
    \end{array} \right.
\end{displaymath}
\end{itemize}
where:
\begin{itemize}
\item $\alpha{\in}$(0,1) is called global pheromone volatility rate representing pheromone persistence, thus $(1-\alpha)$ corresponds
to global evaporation rate of the pheromone trails~{\color{blue}\cite{cai01}}.
\item $\Delta\tau_{ij}$ denotes pheromone increment on edge $(i,j)$ that ant $k$ travelled in current cycle.

When the ACS operates a main loop, the ants start to construct their solutions. Thereafter pheromone trails are updated.
\end{itemize}
Let:
\begin{equation}
g(s)=\frac{Q_g}{{\textnormal{cost}}(p_k(s,u))}
\label{eqn:72}
\end{equation}

where:
\begin{itemize}
\item $\textnormal{cost}(p_k(s,u))$ represents the total costs of path $p_k(s,u)$ which the $k^{th}$ ant chooses.
\item $Q_g$ is an adjustment factor of the global update parameters which may affect global convergence speed of the ACO algorithm.
\item $\rho$ and $\lambda$ should be set up to be smaller than 1 for precluding the pheromone trails from
infinitely accumulating. 
\end{itemize}

\begin{itemize}
\item[4)]{\bf Elitist~strategy:}                              
The pheromone trails deposited on the solutions constructed at this iteration can be reinforced using Elitist strategy~{\color{blue}\cite{afshar02}}. The pheromone trails deposited on the current edge with the optimal objective function found at this iteration is reinforced to facilitate the best solution search around the specified point at the end of every search cycle. Actually, \emph{the elitist~strategy} is used to preserve the optimal individuals from one generation to another~(Elitist Preservation Strategy-EPS) so that the system does not lose the optimal individuals found in the optimization process. The elitist strategy of the ACO algorithm in this paper can
be used to directly copy the best individual at each iteration into the population for the next generation. In the current work, we have investigated the performances of the EACO algorithm with different~{\emph{elitist}} sizes, ~{\emph{population}} sizes, and {\emph{generation numbers}} for improving the global convergent property of the EACO
considering various constrained problems.
\begin{equation}
\tau_{ij}(t+1)=\rho\tau_{ij}(t)+\Delta\tau_{ij}+\Delta\tau^*_{ij}.
\label{eqn:73}
\end{equation}

where:
\begin{displaymath}
\Delta\tau_{ij}=\sum\limits^m_{k=1}\Delta\tau^k_{ij}.
\end{displaymath}
\setlength{\arraycolsep}{0.0em}
\begin{eqnarray}
\Delta\tau^*_{ij}(t)\!=\!\left\{\begin{array}{ll}
\sigma\frac{Q}{f({S_{\textnormal{best}}(t)})} & \textrm{~if~$e_{ij}\!\in\!$ global best solution}          \\
~~0                      & \textrm{~otherwise}
    \end{array} \right.
\label{eqn:74}
\end{eqnarray}
\end{itemize}
where:
\begin{itemize}
\item $\Delta\tau^*_{ij}(t)$ denotes the pheromone increment deposited on the edge~($i,j$) that elitist ants travel.
\item $\sigma$ refers to the number of elitist ants.
\item $Q$ is called a
pheromone reward factor and can quantify the influence of updating information relative to $\tau_0$.
Therefore $Q$ and $\tau_0$ are referred to as dependent parameters. Then, how to define the dependent parameters and the dependence proof
of $Q$ and $\tau_0$, will be described in {\color{blue}Appendices}.
\item $S_{\textnormal{best}}(t)$ represents the set of edges of the best solutions found within iteration $t$.
\end{itemize}

The main characteristics of the elitist ant colony optimization systems are as follows:
\begin{itemize}
\item Elitist~strategy enables the ACO systems with genetic mutation to find best solutions;
\item The elitist ant colony systems with genetic mutation can find best solutions taking the shortest time;
\item Too many elitist ants may result in the premature convergence to sub-optimal solutions in solution searching process.
\end{itemize}
\begin{itemize}
\item[5)]{\bf Termination test:} If the test passed, stop; otherwise, go to local~update.
\end{itemize}
%
\subsection{Gait Optimization for Bipedal Locomotion using EACO}
Main objective of gait optimization and control is to adapt robot's joint angle trajectories like those of the humans. One of the key challenges in robotic bipedal locomotion is to find gait parameters that optimize walking speed, robustness or energy efficiency~{\color{blue}\cite{calandra01}}.
In this article,, we define the fitness value using the distance covered by the humanoid robot in the
expected walking direction during the period of 20 seconds.
\begin{myDef}
\textnormal{\small If the distance covered during the period of the robot falling down is denoted as $l_{\textnormal{dis}}$, when $l_{\textnormal{dis}}$$>$10~(cm), we take the distance value as the fitness value, or this robot obtains a fitness value of 0.}
\end{myDef}
\begin{myDef}
\textnormal{\small If the tested walking parameters of a humanoid robot does not lead to any motion
into the expected walking orientation, then the system rewards the robot with a fitness value of 0.}
\end{myDef}
\begin{myDef}
\textnormal{\small When the deviation from the expected walking orientation of a humanoid robot caused by \emph{drifting} is more than 45
degree, the system offers this robot a fitness value of 0.}
\end{myDef}
For enhancing search capability and convergence rate of the EACO algorithms, the EACO algorithms limit the search space
by choosing the minimum and maximum values of each dimension based on knowledge on the humanoid robot platform, walking trajectories
of the robotics, and their synchronization~{\color{blue}\cite{niehaus02}}.

\subsection{Bipedal Walking Inspired from Humans}
The human learning ability can influence greatly the ability for human's biped walking locomotion~{\color{blue}\cite{salter01,srinivasan01}}.
However, all the robots do not have human-like learning capability. Therefore they are not good at learning from the unknown
complex environments. To develop robots' biped walking locomotion, at first we must investigate robust locomotion strategy and
understand human's biped walking patterns and focus on development of theoretical strategy and computational brain controller
of humanoid robots such that the correct joint trajectories synthesis on the articulated chained manipulators
at the robot's legs {\color{blue}\cite{yussof01,morimoto01}}. Yussof {textit{et al}}. introduced a control strategy of a dynamic
biped walking robot for compensating for the three-axis ({\color{blue}{pitch, roll, and yaw}}) moment on an arbitrary planned ZMP based
on trunk motion and developed a biped walking robot and performed a biped walking experiment with
small-size human robotics based on the control strategy {\color{blue}\cite{yussof02}}.

\subsection{Bipedal Walking Parameters' Optimization}
It is well known that in biped walking locomotion, the kinematical solutions and optimal gait trajectory patterns for humanoid robotics legs should be developed~{\color{blue}\cite{spitz01}}. In this article, we usually construct one cycle by humanoid two steps and the walking speed of biped robot locomotion is controlled by three parameters: step length(s),~hip-joint~height(h)
from the ground, and {duty-ratio(d)}~{\color{blue}\cite{yussof03}}. The order of the robots' joints from hip to foot is regarded
as standard for humanoid robotics, that is hip {\color{blue}{yaw}}, hip {\color{blue}{roll}},
hip {\color{blue}{pitch}}, knee {\color{blue}{pitch}}, ankle {\color{blue}{pitch}}, ankle {\color{blue}{roll}},
and toe {\color{blue}{pitch}}~{\color{blue}\cite{zorjan01}}. The relationship between the optimized joint angles of walking
humanoid robotics and time is illustrated in Fig.~{\color{red}{\ref{efig03}}}({\bf A}).

\subsection{Data Analysis} All experiments were repeated at least five times unless
otherwise stated.

\section*{APPENDIX}
\section*{Proof of Dependence of $Q$ and $\tau_0$}
$Q$ and $\tau_0$ are proved to be dependent parameters based on theoretical analysis of the influence of the pheromone reward factor $Q$ on
the decision-making.
%
\begin{myDef}
\textnormal{\small Without loss of generality, dependence between $Q$ and $\tau_0$ can be defined as the capability to express either one of the parameters as some function of the other parameter, that is, they are referred to as dependent parameters~\color{blue}\cite{zecchin01}.}
\end{myDef}

This implies that one of the dependent parameters can assume any value and the other parameter
can be calibrated to it maintaining performance of the algorithms. As long as one of the parameters
is calibrated, the algorithm's performance is independent of the value that the other dependent
parameter takes.
\begin{lemma}
$Q$ and $\tau_0$ are dependent parameters.
\end{lemma}
\newtheorem{proof}{Proof}
\begin{proof}
\textnormal{Without loss of generality, assuming that the ant $k$ looks for an optimal solution $S_{\mathrm{optimal}}$ in its tour,
and then we consider the following cases. First, we can regard the EACO algorithms as an evolution of time of the probability
functions at each decision-making point. This can be proved that the unique purpose of EACO focuses on changing
the probability functions for constructing and selecting the global best solution most likely. Accordingly, assume that $p_{i,j}(t)$ denotes
the probability that edge $(i,j)$ is selected at iteration $t$, $\tau_{i,j}(t)$ refers to the concentration of pheromone related to edge $(i,j)$
at iteration $t$, we derive an expression for $p_{i,j}(t)$ as a function of $Q$ and $\tau_0$,
but first and foremost we are required to express $\tau_{i,j}(t)$ as a function of $Q$ and $\tau_0$ {\color{blue}\cite{zecchin01}}.}
\end{proof}
From Eq.~[{\color{red}\ref{eqn:32}}], $\tau_{i,j}(t)$ can be expressed as a function of $\tau_{i,j}(t-2)$, $\Delta\tau_{i,j}(t-2)$,
and $\Delta\tau_{i,j}(t-1)$ by
\begin{align}
\tau_{i,j}(t)&=\rho\tau_{i,j}(t-1)+\Delta\tau_{i,j}(t-1)                                                 \nonumber \\
             &=\rho(\rho\tau_{i,j}(t-2)+\Delta\tau_{i,j}(t-2))+\Delta\tau_{i,j}(t-1)                     \nonumber \\
             &=\rho^2\tau_{i,j}(t-2)+\rho\Delta\tau_{i,j}(t-2)+\Delta\tau_{i,j}(t-1)
\label{eqn:08}
\end{align}
In a similar recursive pattern, $\tau_{i,j}(t)$ can be expressed as a function of the pheromone value
in the first iteration and all the pheromone additions up to iteration $(t-1)$. Recognizing
that $\tau_{i,j}(1)=\tau_0$, by recursion the following expression is obtained:
\begin{eqnarray}
\tau_{ij}(t)=\rho^{(t-1)}\tau(0)+\sum_{l=1}^{t-1}\rho^{t-1-l}\Delta\tau_{i,j}(l)
\label{eqn:30}
\end{eqnarray}
To include the $Q$ term, Eq.~[{\color{red}\ref{eqn:32}}] is substituted into Eq.~[{\color{red}\ref{eqn:30}}] to obtain the following expression:
\begin{eqnarray}
\tau_{ij}(t)=\rho^{(t-1)}\tau(0)+\sum_{l=1}^{t-1}\rho^{t-1-l}\delta_{i,j}(l)\frac{Q}{f(S_{\mathrm{best}}(l))}
\label{eqn:31}
\end{eqnarray}

where $\delta_{i,j}(t)$ is a delta function given by
\begin{displaymath}
\delta_{ij}(t)=\left\{\begin{array}{ll}
1 & \textrm{if~an edge $e_{ij}\in$$S_{\mathrm{best}}(t)$.}          \\
0 & \textrm{otherwise}
    \end{array} \right.
\end{displaymath}
This means that edge $(i,j)$ receives the pheromone addition of $Q/f(S_{\mathrm{best}}(t))$ if it is an
element of $S_{\mathrm{best}}(t)$. Substituting Eq.~[{\color{red}\ref{eqn:31}}] into Eq.~[{\color{red}\ref{eqn:02}}], an expression for the evolution of the probability distributions based on the initial pheromone value and the pheromone additions from iteration 1 to $(t-1)$ is achieved. This can be written as
\begin{equation}
p_{i,j}(t)=\frac{\Big[\frac{\tau_0}{Q}+\sum\limits_{l=1}^{t-1}{\frac{\delta_{i,j}(l)}{\rho^lf({S_{\mathrm{best}}(l))}}\Big]^\alpha}
[\eta_{i,j}]^\beta}{\sum\limits_{k{\notin}\mathrm{tabu}_k}\Big[\frac{\tau_0}{Q}+\sum\limits_{l=1}^{t-1}{\frac{\delta_{i,j}(l)}
{\rho^lf({S_{\mathrm{best}}(l))}}\Big]^\alpha}[\eta_{i,j}]^\beta}.
\label{eqn:32}
\end{equation}
The parameters $\tau_0$ and $Q$ appear only together as a ratio in Eq.~[{\color{red}\ref{eqn:32}}], and as Eq.~[{\color{red}\ref{eqn:32}}] represents
a complete expression of the algorithmic
process it can also be seen that $\tau_0$ and $Q$ can be replaced by another parameter (say $\kappa$, where $\kappa=\tau_0/Q$) without loss of generality
in the process. A result of this is that $\tau_0$ can be expressed as a function of $Q$~(i.e.,~$\kappa=\tau_0/Q$) without loss of generality. This
implies that $\tau_0$ and $Q$ are dependent parameters.
\section*{\color{blue}Experiential Values of Parameter Setup}
Experiential values of the parameters setup of the EACO algorithms applied to path planning
and gait optimization are illustrated in Table~{\color{red}{\ref{tab04}}}. The investigated results show
that the parameters options have a great influence upon the performance of the EACO algorithms,
for example, a small population size and relatively large mutation rate is far
superior to the large population sizes and a smaller mutation rate that are used by most of
the papers presented in the combinatorial optimization problems.
%
\section*{ACKNOWLEDGMENTS}
We thank AI Lab. at the Stanford University for experimental and technical support. We thank Professor Auke Ijspeert at the EPFL for helpful suggestions and insightful discussions that greatly improved the manuscript. {\color{blue}\bf{Funding}}: This work is supported in part
by the Natural Science Foundation of China Grants no. 69985003 and no. 61475027 and International Science and Technology Cooperation Program of Changzhou under Grant no.CZ20110016. {\color{blue}\bf{Author contributions}}: J.Y. conceptualized the problem and the technical framework, developed and tested algorithm, designed and performed all experiments, collected and analyzed data, interpreted the results, created pictures and videos, and wrote and edited the paper; J.Y. and V.P. managed the project. All authors discussed the results, commented on the manuscript, and approved the final version of this manuscript. These authors contributed equally to this work.
{\color{blue}\bf{Competing interests}}: The authors declare that they have no competing financial interests.
{\color{blue}\bf{Data and Materials}}: All data needed to evaluate the conclusions in the paper are included in the supplementary materials.

\begin{description}
\item[Data and Materials:] All data needed to evaluate the conclusions in the paper are
present in the paper and/or the Supplementary Materials.
\end{description}

\begin{description}
\item[Authors Information:] The authors declare no competing financial interests. Correspondence should be addressed to J.~A.~Yang (yja431008@gmail.com).
\end{description}
\bibliographystyle{ACM-Reference-Format}
\bibliography{rob.bbl}
%

\begin{table*}[!t]
\renewcommand\arraystretch{1.1}
\centering
\small
\advance\tabcolsep-2.8pt
\caption{Comparison between computing efficiencies of EACO, PSO, GA and SA}
\vspace{-0.3cm}
\begin{threeparttable}
\begin{tabular}{c|ccc} \hline\toprule
\diagbox{Algorithms}{Results}{Experiments} & \tabincell{c|}{Average CPU time(s)\\per iteration} & \tabincell{c|}{Average number of iterations\\needed for convergence} & \tabincell{c}{Average CPU time(s) \\ for optimal solution}                  \\ \hline
{\color{red}EACO}  & {\color{red}0.00048} & {\color{red}117}  & {\color{red}0.0793}        \\
ACO                & 0.00059  & 175 & 0.1033                                               \\
PSO                & 0.00089  & 327 & 0.2489                                               \\
Real-coded GA\color{red}\footnote[1]      & 0.00067  & 912 & 0.6110                        \\
Simulated annealing& 0.00083  & 987 & 0.5873                                               \\ \bottomrule
\end{tabular}
\begin{tablenotes}
\item[{\color{red}1}]Real-coded genetic algorithms with Laplace crossover and power mutation operators make use of real-coded chromosomes to initialize population for optimizing humanoid walking parameters.
\end{tablenotes}
\end{threeparttable}
\label{tab01}
\end{table*}
\begin{table*}[!t]
\renewcommand\arraystretch{1.1}
\small
\centering
\caption{Comparison between the performances of EACO, GA, and SA}
\vspace*{-0.3cm}
\tabcolsep 6pt
\begin{tabular}{c|ccc|ccc|ccc} \hline \toprule
\specialrule{0em}{-0.8pt}{1pt}
 \multirow{2}*{Experiments} & \multicolumn{3}{c|}{Iteration counts} & \multicolumn{3}{c|}{Times to get optimal solutions} & \multicolumn{3}{c}{Average optimal solutions}  \\ \cline{2-4}\cline{5-7}\cline{8-10} & GA  &  SA  &{\color{red}EACO}  &  GA  & SA  & EACO  & GA  & SA  &{\color{blue}EACO}  \quad \\ \hline
  $G_1$ & 894  & 598  &{\color{red}472} &60.08   & 53.86  & 40.76 &$-$14.136 & $-$14.571    &{\color{blue}$-$14.957} \quad \\
  $G_2$ & 4918 & 3067 &{\color{red}2007}&352.39  & 205.24 & 156.03 &8063.12  & 7128.06      &{\color{blue}~7051.12}  \quad \\
  $G_3$ & 5828 & 4679 &{\color{red}4182}&451.73  & 384.12 & 342.54 &682.114  &~687.293      &{\color{blue}~680.203}  \quad \\
  $G_4$ & 5301 & 4023 &{\color{red}3894}&411.48  & 365.40 & 319.84 & 0.056   &~0.055        &{\color{blue}~0.05578}  \quad \\
  \bottomrule
  \end{tabular}
  \label{tab02}
\end{table*}

\vspace*{-1.6cm}
\begin{table*}[!t]
\renewcommand\arraystretch{1.1}
\small
\centering
\caption{Influence of the different parameters on the experimental results.}
\vspace{-0.3cm}
\tabcolsep 10pt
       {\begin{tabular}{@{}cccc|cc@{}} \hline\toprule
       \specialrule{0em}{-0.8pt}{1pt}
       \multicolumn{4}{c|}{\qquad Parameter setup} &\multicolumn{2}{c}{\qquad Iteration counts and optimal solutions\qquad\qquad } \\  \cline{1-6}
       \qquad $q_0$\quad~~~& $\rho$ & $p_{\rm{crossover}}$ & $p_{\rm{mutation}}$ &\quad Average iteration counts & \quad Optimal solutions\qquad\qquad\\ \hline
       \qquad 0.2\quad~~~&0.1& 0.2 & 0.2 & 599.2 & ~~~~$-$14.957   \qquad \\
       \qquad 0.4\quad~~~&0.2& 0.3 & 0.3 & 521.4 & ~~~~$-$14.951   \qquad \\
       \qquad 0.6\quad~~~&0.3& 0.4 & 0.4 & 544.8 & ~~~~$-$14.969   \qquad \\
       \qquad {\color{red}0.8~~}\quad~&{\color{red}0.3}& {\color{red}0.6} &{\color{red}0.5}&{\color{red}393.6}&~~~~{\color{red}$-$14.952} \qquad\\
       \qquad 0.9\quad~~~&0.4& 0.8 & 0.8 & 639.7 & ~~~~$-$14.993  \qquad\\ \bottomrule
    \end{tabular}}
    \label{tab03}
\end{table*}
\begin{table*}[!t]
\renewcommand\arraystretch{1.1}
\centering
\small
\caption{Experiential values of the EACO parameters setup for path planning and gait optimization issues in this article.}
\vspace{-0.3cm}
\begin{threeparttable}
\tabcolsep 11pt
{\begin{tabular}{@{}c||ccc@{}} \hline\toprule
\quad Parameters & Description & Actual values  & Values/expression                                                   \quad\qquad  \\ \hline
\quad $\tau_0$   & Initial pheromone value      & 0.05 & $\frac{1}{n{\cdot}L_{nn}}$\color{red}\footnote[2]            \quad\qquad \\
\quad $\alpha$   & Decision control parameters for pheromone           & 1.0  & $0.5-5.0$                             \quad\qquad  \\
\quad $\beta$    & Decision control parameters for desirability        & 5.0  & $1.0-5.0$                             \quad\qquad  \\
\quad $\rho$     & Pheromone persistence factor                        & 0.2  & $0.1-0.9$                             \quad\qquad  \\
\quad $\eta$     & Desirability of edges                               & 0.3  & $0.1-0.3$                             \quad\qquad  \\ \hline
\quad $m$        & Number of ants                                      & 20,40& $20-60$                               \quad\qquad  \\
\quad $Q$        & Pheromone reward factor & 100 & $C(\Omega^{max})$\color{red}\footnote[3]                           \quad\qquad  \\
\quad $PEN$      & Penalty factor                                      & 0.3  & $0.3-0.8$                             \quad\qquad  \\ \hline
\quad $p_{\mathrm{mutation}}$ & Mutation rate for genetic algorithms   & 0.5  & $0.05-0.5$                            \quad\qquad  \\
\quad $p_{\mathrm{crossover}}$& Crossover probability for GAs          & 0.6  & $0.5-1.0$                             \quad\qquad  \\ \bottomrule
\end{tabular}}
\begin{tablenotes}
\item[{\color{red}2}]where $n$ is the number of nodes and
$L_{nn}$ is the shortest path length between neighbor nodes.
\item[{\color{red}3}]$Q$ can be calculated based on $C(\Omega^{max})$ proposed by Zecchin \emph{et al}.{\color{blue}\cite{zecchin01}}.
\end{tablenotes}
\end{threeparttable}
\label{tab04}
\end{table*}
\end{document}